\documentclass{article}

\usepackage[accepted]{icml2026}

\usepackage[utf8]{inputenc} % allow utf-8 input
\usepackage[T1]{fontenc}    % use 8-bit T1 fonts
\usepackage{hyperref}       % hyperlinks
\usepackage{url}            % simple URL typesetting
\usepackage{booktabs}       % professional-quality tables
\usepackage{amsfonts}       % blackboard math symbols
\usepackage{nicefrac}       % compact symbols for 1/2, etc.
\usepackage{microtype}      % microtypography
\usepackage{xcolor}
\definecolor{mydarkblue}{rgb}{0,0.08,0.45}
\definecolor{dmorange500}{HTML}{FF5F19}
\definecolor{dmblue300}{HTML}{2267EB}
\definecolor{dmred300}{HTML}{FF617B}

\hypersetup{
	colorlinks,
	citecolor=dmblue300,
	linkcolor=dmred300,
	urlcolor=mydarkblue}

\usepackage{amsmath}
\usepackage{amssymb}
\usepackage{mathtools}
\usepackage{amsthm}
\usepackage{wrapfig}
\usepackage{graphicx}
\usepackage{subcaption}
\usepackage{dsfont} % for indicators

\usepackage{enumitem}
\usepackage{newtxmath}

\usepackage[table]{xcolor}
\definecolor{bestcolor}{gray}{0.85}

\usepackage[capitalize,noabbrev]{cleveref}

\theoremstyle{plain}
\newtheorem{theorem}{Theorem}[section]
\newtheorem{proposition}[theorem]{Proposition}

\newtheorem{corollary}[theorem]{Corollary}
\theoremstyle{definition}
\newtheorem{definition}[theorem]{Definition}
\newtheorem{assumption}[theorem]{Assumption}
\theoremstyle{remark}
\newtheorem{remark}[theorem]{Remark}

\usepackage[textsize=tiny]{todonotes}

\crefname{section}{\S\@gobble}{\S\@gobble}
\crefname{subsection}{\S\@gobble}{\S\@gobble}
\crefname{definition}{Def.}{Defs.}
\crefname{Proposition}{Prop.}{Props.}

\AddToHook{env/assumption/begin}{\crefalias{theorem}{assumption}}
\AddToHook{env/definition/begin}{\crefalias{theorem}{definition}}
\AddToHook{env/proposition/begin}{\crefalias{theorem}{proposition}}
\AddToHook{env/corollary/begin}{\crefalias{theorem}{corollary}}

\makeatletter
\crefname{section}{\S\@gobble}{\S\@gobble}
\crefname{subsection}{\S\@gobble}{\S\@gobble}
\crefname{definition}{Def.}{Defs.}
\crefname{Proposition}{Prop.}{Props.}

\AddToHook{env/assumption/begin}{\crefalias{theorem}{assumption}}
\AddToHook{env/definition/begin}{\crefalias{theorem}{definition}}
\AddToHook{env/proposition/begin}{\crefalias{theorem}{proposition}}
\AddToHook{env/corollary/begin}{\crefalias{theorem}{corollary}}

\renewcommand{\section}{%
  \@startsection{section}{1}{\z@}%
                {-1.0ex \@plus -0.2ex \@minus -0.2ex}%
                { 1.0ex \@plus  0.2ex \@minus  0.2ex}%
                {\large\bf\raggedright}%
}
\renewcommand{\subsection}{%
  \@startsection{subsection}{2}{\z@}%
                {-0.8ex \@plus -0.2ex \@minus -0.2ex}%
                { 0.8ex \@plus  0.2ex \@minus -0.2ex}%
                {\normalsize\bf\raggedright}%
}
\renewcommand{\subsubsection}{%
  \@startsection{subsubsection}{3}{\z@}%
                {-0.6ex \@plus -0.2ex \@minus -0.2ex}%
                { 0.5ex \@plus  0.2ex \@minus -0.2ex}%
                {\normalsize\bf\raggedright}%
}
\renewcommand{\paragraph}{%
  \@startsection{paragraph}{4}{\z@}%
                {0ex}%
                {-1em}%
                {\normalsize\bf}%
}

\makeatother

\icmltitlerunning{Stop the Sampler! Classifier-Based Adaptive Stopping for Sampling Kernels}

\begin{document}

\twocolumn[
  \icmltitle{Stop the Sampler! Classifier-Based Adaptive Stopping for Sampling Kernels}



  \icmlsetsymbol{equal}{*}

  \begin{icmlauthorlist}
    \icmlauthor{Kirill Korolev}{hse}
    \icmlauthor{Nikita Morozov}{hse}
    \icmlauthor{Stepan Pavlenko}{hse}
    \icmlauthor{Esmeralda S. Whitammer}{ue,cifar}
    \icmlauthor{Sergey Samsonov}{hse}
  \end{icmlauthorlist}

  \icmlaffiliation{hse}{HSE University}
  \icmlaffiliation{ue}{University of Edinburgh}
  \icmlaffiliation{cifar}{CIFAR Fellow}

  \icmlcorrespondingauthor{Kirill Korolev}{kkorolev@hse.ru}

  \icmlkeywords{Machine Learning, ICML}

  \vskip 0.3in
]



\printAffiliationsAndNotice{}  % no special notice (required even if empty)
\begin{abstract}
  Sampling from complex, unnormalized probability densities is a fundamental challenge in Bayesian inference and probabilistic modeling. While Markov chain Monte Carlo (MCMC) methods provide asymptotic guarantees, they often suffer from slow mixing and high computational costs due to fixed or manually tuned trajectory lengths. In this work, we propose a novel framework that treats trajectory termination as a learnable component of the sampling dynamics. By framing MCMC within the theory of non-acyclic generative flow networks (GFlowNets), we train state-dependent neural classifiers to decide when a trajectory has reached a high-density region and should terminate. We theoretically establish the connection between optimal classifiers and the target density via detailed balance conditions and introduce a multilevel training scheme to facilitate exploration in complex geometries. Experimental results across various benchmark densities demonstrate that our approach significantly reduces average trajectory lengths while improving mode coverage and mixing compared to standard MCMC baselines.
\end{abstract}

\section{Introduction}
\label{sec:intro}
In this paper, we consider the problem of sampling from a probability distribution given by its unnormalized density:
\begin{equation}
\label{eq:sample_problem}
\textstyle 
\pi(x) = r(x)/Z, \quad Z = \int_{\mathbb{R}^d} r(x) \,dx 
\end{equation}
where $r: \mathbb{R}^d \rightarrow \mathbb{R}^{+}$ is a positive function and $Z$ is an unknown normalizing constant. The problem arises in many domains, including Bayesian statistics, probabilistic deep learning, and natural sciences~\citep{liu2001monte, stoltz2010free, welling2011bayesian, noe2019boltzmann, izmailov2021bayesian}. The main two approaches to solve the sampling problem \eqref{eq:sample_problem} are either to perform (asymptotically) exact sampling from $\pi$ based on Monte Carlo or MCMC algorithms \cite{neal2001annealed,neal2011mcmc,hoffman2014no}, or to bypass exact sampling in favor of \emph{amortized} solutions to this problem. MCMC algorithms operate by constructing an ergodic Markov chain (see, e.g., \citet{douc:moulines:priouret:soulier:2018}) $\{Y_n\}_{n \in \mathbb{N}}$ with invariant distribution equal to $\pi$, such that the distribution of $n$-th iterate $Y_n$ becomes close to $\pi$ regardless of the starting distribution of $Y_0$. MCMC-based approaches are known to suffer from slow convergence to the target distribution \cite{andrieu2003introduction,neal2011mcmc}, especially when the target dimension $d$ is large. In addition to slow convergence, the important question related to this family of methods is whether the length of the generated trajectory is sufficient for the marginal distribution of the last element to be close to $\pi$. Known approaches to this problem consider some statistics based on generated samples \cite{mcmcdiag2019,mcstopreview2025}, which do not always provide affirmative stopping criteria. 

Contrary to exact sampling approaches, amortized solutions to \eqref{eq:sample_problem} train a generative model to approximately sample from $\pi$. Contrary to MCMC, they shift the computational burden to the training phase, enabling faster inference at the price of bypassing the exact sampling. Models of this class include normalizing flows~\cite{noe2019boltzmann, midgley2023flow,gabrie2022adaptive}, diffusion samplers \cite{zhang2021path,vargas2023denoising,blessing2024beyond}, and generative flow networks (GFlowNets) \cite{bengio2021flow, bengio2023gflownet, lahlou2023theory}. While GFlowNets were initially developed for discrete domains, they were further extended to continuous problems in \citet{lahlou2023theory} and offer another family of training algorithms for diffusion samplers \citep{zhang2023diffusion,sendera2024improved,kim2025scalable,gritsaev2025adaptive,berner2026from}. GFlowNets operate by training a stochastic policy to approximate the target distribution over terminal states in an appropriately constructed Markov decision process (MDP), see, e.g., \citet{tiapkin2024generative}. Recent theoretical developments in GFlowNets \cite{lahlou2023theory,deleu2023generative,brunswic2024theory} present a broad and convenient theoretical framework for developing sampling algorithms.

In this work, we utilize GFlowNet methodology to train neural network classifiers for early stopping of MCMC kernels, creating a new family of amortized samplers based on algorithms from both worlds. Our key idea is to treat stopping as a learnable component of the sampling dynamics: instead of running a Markov chain for a fixed or externally determined number of steps, we learn state-dependent classifiers that decide when a trajectory should terminate. This leads to a unified framework where sampling trajectories are dynamically shortened in high-density regions while remaining exploratory elsewhere. Our main contributions are:

\begin{enumerate}[left=0pt,nosep]
    \item We theoretically show how sampling dynamics in $\mathbb{R}^d$ with an addition of a possibility to stop in any point $x$ can be viewed as a non-acyclic GFlowNet environment~\cite{brunswic2024theory}, establish connections between the learned stopping policies and the target density via flow and detailed balance conditions, and derive a characterization of optimal classifiers in terms of underlying MCMC dynamics. 
    \item Building on this foundation and prior non-acyclic GFlowNet training methodology~\cite{brunswic2024theory, morozov2026learning}, we introduce practical algorithms that combine learned stopping with adaptive forward and backward kernels, as well as a multilevel extension that improves training and exploration. 
    \item We provide experimental evaluation on a number of density functions, demonstrating that our approach yields substantially shorter trajectories compared to standard MCMC while better capturing the geometry of complex target distributions and improving mixing across modes through learned drift corrections.
\end{enumerate}
Source code: \href{https://github.com/kkorolev1/stop-the-sampler}{github.com/kkorolev1/stop-the-sampler}.

\section{Background}
\subsection{GFlowNets on general state spaces}

We briefly describe the setting of the sampling problem on the general state space. Given a measurable space $(\mathcal X, \Sigma_{\mathcal{X}})$, and a finite reward measure $R$ on this space, we aim to generate samples from the distribution $\pi$ defined by
\[
%\textstyle 
\pi(A) = {R(A)}/{R(\mathcal X)},\quad A \in \Sigma_{\mathcal{X}}\,.
\]
Following \citet{lahlou2023theory}, we solve this problem with an extension of GFlowNets, which can be defined on general state spaces using a measure-theoretic extension of directed acyclic graphs. This structure is represented by a \emph{measurable pointed graph} $\mathcal G = (\bar{\mathcal{S}}, \Sigma, s_0, s_f, \kappa_F, \kappa_B, \nu)$. Here, $(\bar{\mathcal{S}}, \Sigma)$ is a measurable space with $\bar{\mathcal{S}} = \{s_0\} \cup \mathcal{S} \cup \{s_f\}$, where $s_0,s_f\in\bar{\mathcal{S}}$ are the distinguished source and sink states. We assume that $\mathcal X = \{s \in \mathcal{S} : \kappa_F(s,\{s_f\}) > 0\}$ denote the set of terminal states, and $\nu$ is some reference measure on $\mathcal{S}$, such that $R \ll \nu$. In this statement, $\kappa_F(s,\cdot)$, and  $\kappa_B(s,\cdot)$ are the reference forward and backward kernels. The supports of $\kappa_F(s,\cdot)$ and $\kappa_B(s,\cdot)$ generalize the notions of children and parents in a discrete DAG, respectively, defining the set of states reachable from $s$ (resp.\ from which $s$ is reachable) in one step. The kernels $\kappa_F$ and $\kappa_B$ must satisfy several compatibility conditions, in particular, it should exist $N \in \mathbb{N}$, such that 
\begin{equation}
\label{eq:finite_traj_condition}
\textstyle
\operatorname{supp}(\kappa_F^{N}(s_0,\cdot)) = \{s_f\}\,.
\end{equation}

\textbf{Trajectory distribution and sampling problem.} Suppose that we are given a \emph{forward} Markov kernel $P_F$ on $(\bar{\mathcal{S}}, \Sigma)$ such that $P_F(s,\cdot) \ll \kappa_F(s,\cdot)$. Learner should be able to sample from $P_F(s,\cdot)$. The kernel $P_F(s,\cdot)$ naturally induces a measure over $(\mathcal{S}^{n},\Sigma^{\otimes n})$ given by 
\begin{equation}
\textstyle
P_F^{\otimes n}(s_0,\mathrm{d} s_{1:n}) = \prod_{i=1}^{n} P_F(s_{i-1}, \mathrm{d} s_i)\,.
\end{equation}
Under appropriate regularity conditions (see \citet{lahlou2023theory}), $P_F^{\otimes n}$ induces the probability distribution over the complete trajectories $\tau = (s_0 \to s_1 \to \dots \to s_{n} \to s_{n+1}=s_f)$ of all possible length $n \in \mathbb{N}$ given for any $C_n \in \Sigma^{n}$ by  
\begin{equation}
\begin{aligned}
\textstyle
&P(\tau \in (\{s_0\} \otimes C_n \otimes \{s_f\})) =\\
&\int_{\bar{\mathcal{S}}^{n+1}} P_F^{\otimes (n+1)}(s_0, \mathrm{d} s_{1:n+1}) \mathbb{I}\{s_{1:n} \in C_n\} \mathbb{I}\{s_{n+1} = s_f\}\,.
\end{aligned}
\end{equation}
Note that the condition \eqref{eq:finite_traj_condition} implies that a trajectory sampled starting from $s_0$ according to the probability distribution $P(\tau)$ defined above will terminate with probability $1$ in the number of steps not exceeding $N$. Marginalization of the above trajectory distribution $P(\tau)$ over the terminal states $s_n$ induces the terminating distribution $P_{\operatorname{T}}(B)$ on $(\mathcal{X},\Sigma_{\mathcal{X}})$. The goal of our sampling procedure is to ensure proper conditions on the forward Markov kernel $P_F$ such that for any $A \in \Sigma_{\mathcal{X}}$
\begin{equation}
\label{eq:P_T_condition}
P_{\operatorname{T}}(A) = \pi(A)\,.
\end{equation}

\textbf{Local balance conditions.} Given a $\sigma$-finite measure $F \ll \nu$ over $(\bar{\mathcal{S}}, \Sigma)$ and a Markov kernel $P_F$, the tuple $(F, P_F)$ is said to satisfy \emph{flow-matching conditions} if for any $A \subseteq \bar{\mathcal{S}} \setminus \{ s_0 \}$:
\begin{equation}
\label{eq:fm}
\textstyle
F(A) = \int_{\bar{\mathcal{S}} \setminus \{ s_f\}} F(\mathrm{d}s)\, P_F(s,A),
\end{equation}
where $F$ and $P_F$ are said to be a flow measure and a forward kernel, respectively, over $\mathcal G$. The tuple $(F, P_F)$ satisfies the \emph{reward-matching conditions} if for any $x \in \mathcal X$:
\begin{equation}
\label{eq:rm}
R(\mathrm{d}x) = F(\mathrm{d}x)\, P_F(x,\{s_f\}).
\end{equation}
Specifically, \citet{lahlou2023theory} showed that when the flow and reward matching conditions are satisfied, 
the terminating distribution $P_{\operatorname{T}}$ induced by $P_F$ is proportional to $R$, and \eqref{eq:P_T_condition} holds.

\textbf{Detailed balance.} Given a backward Markov kernel $P_B$ on $(\bar{\mathcal{S}}, \Sigma)$ with $P_B(s',\cdot) \ll \kappa_B(s',\cdot)$, the tuple $(F, P_F, P_B)$ is said to satisfy \emph{detailed balance conditions} if:
\begin{equation}
\label{eq:db}
F(\mathrm{d}s)\, P_F(s,\mathrm{d}s') = F(\mathrm{d}s')\, P_B(s',\mathrm{d}s).
\end{equation}
Furthermore, \citet{lahlou2023theory} demonstrated that if $(F, P_F, P_B)$ satisfies the detailed balance conditions, then $(F, P_F)$ satisfies the flow-matching conditions. Equations \eqref{eq:rm} and \eqref{eq:db} generalize the familiar equations in the discrete case \cite{bengio2023gflownet} for a reward function $R: \mathcal{X}\to\mathbb{R}^{+}$, a flow function $F:\bar{\mathcal{S}}\to\mathbb{R}_{\geq0}$ and policies $P_F,P_B$ represented as transition probability masses:
\begin{equation}
\begin{aligned}
\label{eq:discrete_local_balance}
    R(x)&=F(x)P_F(s_f\mid x),\\
    F(s)P_F(s'\mid s)&=F(s')P_B(s\mid s').
\end{aligned}
\end{equation}
If one defines $r(s), f(s), p_F(s' \mid s)$ and $p_B(s \mid s')$ as the Radon--Nikodym derivatives of the reward measure $R$, flow measure $F$, forward kernel $P_F$, and backward kernel $P_B$ with respect to the corresponding reference measures and kernels,
then \eqref{eq:rm} and \eqref{eq:db} can be rewritten in terms of densities to resemble \eqref{eq:discrete_local_balance} directly:
\begin{equation}
\begin{aligned}
\label{eq:cont_local_balance}
    r(x)&=f(x)p_F(s_f\mid x),\\
    f(s)p_F(s'\mid s)&=f(s')p_B(s\mid s'),
\end{aligned}
\end{equation}
for $\nu$-a.e. $x \in \mathcal{X}$ and $(\nu\otimes\kappa_F)$-a.e. $(s,s')$.

\textbf{Trajectory balance.} Fix $n \in \mathbb{N}$ and let $\tau = (s_0 \to s_1 \to \dots \to s_{n}=x \to s_{n+1}=s_f)$ denote a complete trajectory of length $n$ starting at $s_0$, passing through non-terminal states in $\mathcal S$, reaching a terminal state $s_n = x \in \mathcal X$, and then transitioning to $s_f$. Given $Z \in \mathbb{R}^{+}$, the tuple $(Z, P_F, P_B)$ is said to satisfy trajectory balance conditions if for any $n \in \mathbb{N}$, 
\begin{equation}
\label{eq:tb}
Z \prod_{t=0}^{n} p_F(s_{t+1} \mid s_t)
=
r(s_n)\prod_{t=0}^{n-1} p_B(s_t \mid s_{t+1}),
\end{equation}
for $\kappa_F^{\otimes n}$-a.e. $(s_0,s_1,\ldots,s_{n},s_f)$. Moreover, \citet{lahlou2023theory} showed that the trajectory balance conditions imply both flow and reward matching. 

Note that the framework described above corresponds to the finitely absorbing setting \eqref{eq:finite_traj_condition}, when the trajectories sampled according to $P_F(s,\cdot)$ are guaranteed to have finite length. Later we show how this setting can be generalized under the \cref{ass:finite_exp_len}, see the detailed discussion in \cref{sec:main_algorithm}.

\subsection{Generalizations to non-acyclic setting}
The theory of non-acyclic GFlowNets was developed in \citet{brunswic2024theory}, extending the standard framework to settings where the graph may contain cycles. In this case, additional considerations arise due to the presence of cycles, which may lead to unnecessarily long trajectories. It was shown that the expected trajectory length is controlled by the total flow through non-terminal states, yielding the bound
\begin{equation}
\mathbb E_{\tau \sim P}[n_\tau] \le \frac{F(\mathcal S)}{F(\{s_0\})}.
\end{equation}
Subsequently, \citet{morozov2025revisiting} refined this result for the discrete case by showing that, when \(F\) corresponds to the occupancy measure induced by the backward process, the inequality becomes an equality. These results highlight that minimizing the total flow \(F(\mathcal S)\) can be an effective strategy for reducing expected trajectory lengths in non-acyclic GFlowNets, providing a principled motivation for incorporating flow regularization into training objectives.

\subsection{Related work on adaptive kernels and early-stopping in MCMC}
Another line of work in Monte Carlo studies stopping rules for Markov chains, i.e., criteria for deciding whether a partial trajectory is sufficient for estimation. These rules are typically based on convergence diagnostics or precision criteria computed from the generated samples \citep{mcmcdiag2019,mcstopreview2025}. For example, \citet{mcmcstopping2023} compare Gelman--Rubin, Geweke diagnostics, and effective sample size, showing that aggressive thresholds may lead to premature termination and biased estimates. Thus, stopping in MCMC is usually an online diagnostic applied to the generated trajectory. We provide the corresponding formulas for these diagnostic approaches in \cref{appendix:mcmc-stopping}.

A complementary direction focuses on adaptive kernels, where the transition mechanism is adjusted according to local geometry or performance. In adaptive MCMC \citep{haario2001adaptive,andrieu2008adaptive,vihola2012robust,gradadaptmcmc2019}, proposal parameters are optimized within families such as Gaussian random-walk or MALA.
Extensions include position-dependent preconditioning and state-dependent step sizes \citep{posdepmala2023,automala2023,autostep2024}. More recent work formulates proposal adaptation as a reinforcement learning problem \citep{wang2024rlmh,wang2025harnessing}.

\section{Non-acyclic GFlowNets with adaptive sampling time}
\subsection{Classifier-based adaptive stopping}
\label{sec:main_algorithm}

\paragraph{Structured state space.}
In the context of continuous acyclic GFlowNets, it is common to encode the state space $\mathcal{S}$ as pairs $(s, n)$, where $s \in \mathbb{R}^d$ and $n$ indicates a discrete timestep along a trajectory \citep{lahlou2023theory}. In contrast, in this work we remove the notion of time, moving instead to the non-acyclic GFlowNet setting.

\begin{figure}[t]
    \centering
    \includegraphics[width=0.49\textwidth]{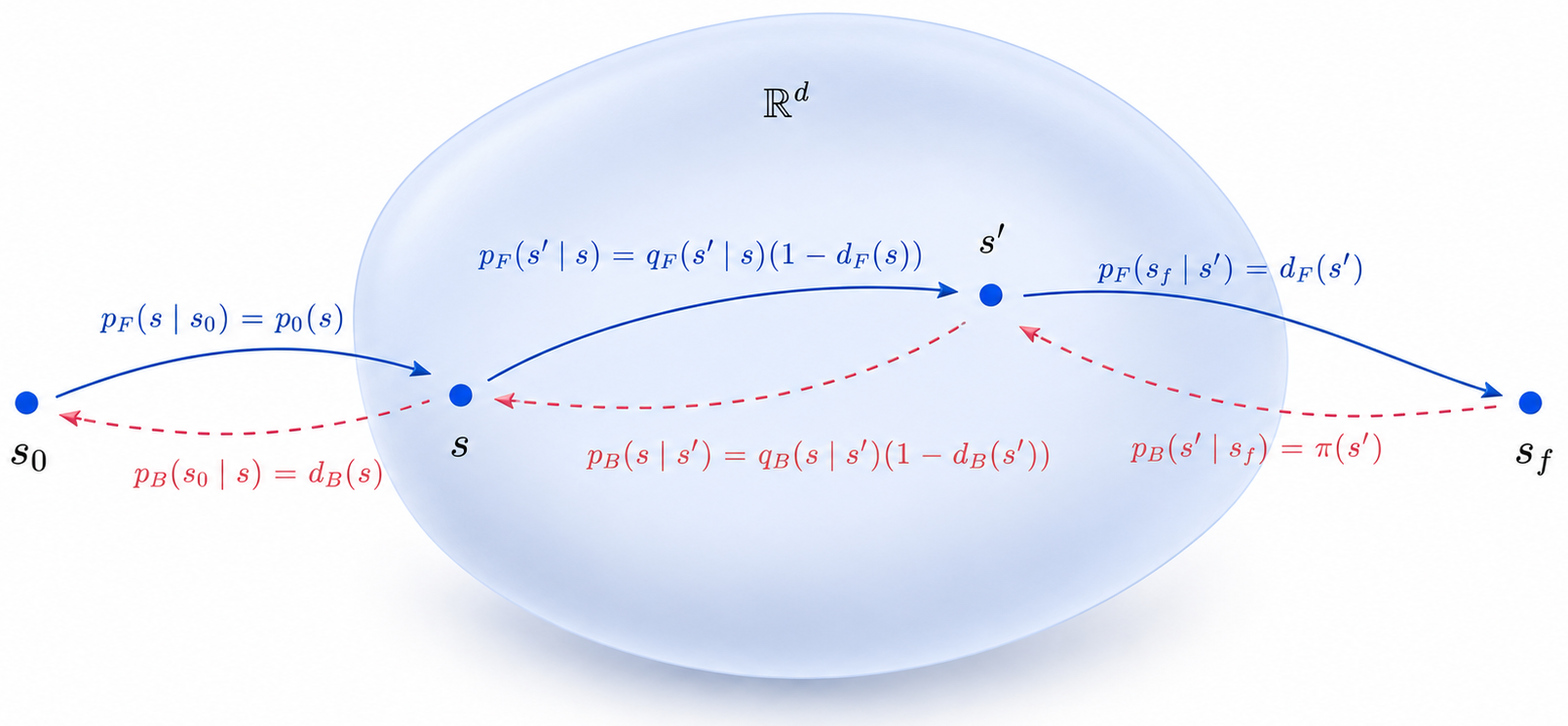}
    \caption{Structured state space for the non-acyclic continuous GFlowNet. The generation process starts from the source state \(s_0\), enters the continuous space \(\mathbb R^d\) according to \(p_0\), evolves between continuous states using \(q_F(s'\mid s)(1-d_F(s))\), and terminates at \(s_f\) with probability \(d_F(s)\). Backward transitions are defined analogously in the reverse direction.}
    \label{fig:main_state_space}
\end{figure}

We define the state space as \(\bar{\mathcal{S}} = \{s_0\} \cup \mathcal S \cup \{s_f\}\), where \(s_0\) and \(s_f\) denote the source and sink states, respectively, and $\mathcal S = \mathbb{R}^d$. Forward transitions are allowed only from \(s_0\) to \(\mathcal S\) and from \(\mathcal S\) to \(\mathcal S\cup\{s_f\}\), and \(s_f\) is absorbing. Backward transitions are defined analogously in the reverse direction. Following \citet{lahlou2023theory}, we model the forward and backward policies as Markov kernels on \(\bar{\mathcal{S}}\). We equip \(\bar{\mathcal S}\) with the reference measure $\nu=\lambda+\delta_{s_0}+\delta_{s_f}$,
where \(\lambda\) is the Lebesgue measure on \(\mathbb R^d\) and \(\delta_s\) is the Dirac measure at \(s\). Throughout, we assume that the relevant measures are absolutely continuous with respect to the reference measure \(\nu\). In turn, a sequence $\tau=(s_0 \to s_1 \to \ldots \to s_{n_{\tau}} \to s_{n_{\tau}+1}=s_f)$ is called a trajectory of length $n_{\tau} \in \mathbb{N}$ if $s_t \in \mathbb{R}^d, t \in \{1,\ldots,n_{\tau}\}$ and $\mathcal T$ denotes the set of such complete trajectories. \cref{fig:main_state_space} illustrates the proposed non-acyclic state space. 

\begin{assumption}
\label{ass:finite_exp_len}
Let \(P\) be the trajectory distribution induced by the forward kernel \(P_F\), then we assume
$\mathbb E_{\tau\sim P}[n_\tau] < \infty$.
\end{assumption}
In particular, \cref{ass:finite_exp_len} is justified and used in all previous non-acyclic works \citep{brunswic2024theory, morozov2025revisiting, morozov2026learning}. In \cref{sec:param} we argue why it holds in our case in practice.
\paragraph{Flow measure and reward matching.}
In contrast to \citet{lahlou2023theory}, where the flow is defined as a measure satisfying the flow-matching condition, we define it directly in terms of the forward transition kernel \(P_F\). This is complementary to the approach of \citet{morozov2025revisiting}, where the flow is constructed via a backward transition kernel \(P_B\).
\begin{definition}[Flow measure]
\label{def:flow}
Under \cref{ass:finite_exp_len}, the flow measure $F$ is defined as
\begin{equation}
\begin{aligned}
\label{eq:flow_def}
F(\{ s_0 \}) &= F(\{ s_f \}) = Z,\\
F(A) &= Z \sum_{n=0}^{\infty} P_F^n(s_0, A), \quad A \in \Sigma_{|\mathcal S}.
\end{aligned}
\end{equation} 
Consequently, as in the discrete case \citep{morozov2025revisiting}, this definition of a flow measure yields flow-matching.
\end{definition}
\begin{proposition}
\label{prop:flow_yiels_fm}
The flow measure $F$ from \cref{def:flow} and the Markov kernel $P_F$ satisfy the flow-matching conditions \eqref{eq:fm}.
\end{proposition}
The proof is in \cref{proof:flow_yiels_fm}. Note that our structured state space is a particular instance of the measurable non-acyclic setting in \citet{brunswic2024theory}. Hence, their general sampling result applies directly in our case.
\begin{theorem}
\label{thm:fm_rm_yields_target}
If a flow measure $F$ satisfies both flow and reward matching conditions, then the sampling distribution $P_T(A)=\pi(A)$.
\end{theorem}
In contrast to \citet{brunswic2024theory}, where the flow is arbitrary and yields only an upper bound on the expected trajectory length, our flow is defined as the occupancy measure induced by \(P_F\), and the bound becomes tight:

\begin{proposition}
\label{prop:exp_traj_len_eq_flow}
Let \(P\) be the trajectory distribution induced by the forward kernel \(P_F\). Let \(F\) be the flow measure (\cref{def:flow}).
Then the expected trajectory length is a constant multiple of the total flow:
\begin{equation} 
\mathbb E_{\tau\sim P}[n_\tau]
=
\tfrac{F(\mathcal S)}{F(\{s_0\})}.
\end{equation}
\end{proposition}
The proof can be found in \cref{proof:exp_traj_len_eq_flow}. To handle trajectory termination without fixed length, we introduce forward and backward classifiers, $d_F, d_B : \mathbb{R}^d \to [0,1]$, which map a state $s \in \mathbb{R}^d$ to the probability of stopping the trajectory. The forward classifier $d_F(s)$ determines the probability of transitioning directly to the sink state $s_f$, while the backward classifier $d_B(s)$ controls the probability of moving to the source state $s_0$. The forward transition density on \(\bar{\mathcal S}\) is defined by
\begin{equation}
\begin{aligned}
p_F(s\mid s_0)&=p_0(s),\\
p_F(s'\mid s)&=(1-d_F(s))q_F(s'\mid s),\\
p_F(s_f\mid s)&=d_F(s),
\end{aligned}
\end{equation}
where $p_0$ is a fixed initial distribution and $q_F$ is a transition density on \(\mathbb R^d\). The backward transition density \(p_B\) on \(\bar{\mathcal S}\) is defined analogously in the reverse direction using transition density $q_B$ on \(\mathbb R^d\) and a classifier $d_B$, with the sink transition fixed by reward matching $p_B(s\mid s_f)=\pi(s)$.

In \cref{proof:derive_flow_from_db}, we show that detailed balance implies the following relations:
\begin{equation}
\begin{aligned}
\label{eq:flow_reward_clf}
    f(s) &= \frac{r(s)}{d_F(s)} = \frac{Z \, p_0(s)}{d_B(s)},\\
    d_F(s) &= \frac{\pi(s)}{p_0(s)} d_B(s).
\end{aligned}
\end{equation}

\subsection{Connection with MCMC}

In this section, we relate the generation kernel restricted to transitions within $\mathcal{S}$ to early stopping of time-homogeneous chains that use this kernel as the their transition operator, which establishes a precise link between MCMC and our framework.

Let $Q_F$ be a fixed forward kernel with a transition density $q_F$. Assume that \(Q_F\) has full support on \(\mathbb R^d\) and is uniformly geometrically ergodic. That is, \(Q_F\) admits a unique stationary measure \(\pi_Q\), and there exist constants \(C<\infty\) and \(\rho\in(0,1)\) such that
\begin{equation}
\label{eq:uge}
    \| \nu Q_F^n - \pi_Q \|_{\mathrm{TV}} \le C\rho^n
\end{equation}
for all \(n\in\mathbb N\) and any probability measure \(\nu\), where \(\|\cdot\|_{\mathrm{TV}}\) denotes the total variation norm. Let $\pi(s) = r(s) / Z$ be a target distribution, $p_0$ be the initial distribution of a Markov chain, and $P$ be the trajectory distribution induced by $p_F=(q_F, d_F)$.  By a slight abuse of notation, we use $p_0, \pi$ and $\pi_Q$ to denote both the corresponding probability measures and their densities, depending on the context.
\begin{theorem}
\label{thm:conn_mcmc}
Assume $Q_F$ is a kernel with a density $q_F$ satisfying~\eqref{eq:uge}. The tuple \((f, q_F, d_F, q_B, d_B)\) is a solution to detailed balance conditions \eqref{eq:db} if and only if the following two statements hold:

1. The flow measure $F$ satisfies for any measurable $A \subseteq \mathbb{R}^d$
\begin{equation}
\label{eq:mcmc_flow_form}
F(A)
=
Z\bigl(U(A)+n_Q\pi_Q(A)\bigr), 
\end{equation}
where
\begin{equation}
\begin{aligned}
U(A)
&=
\sum_{n=0}^{\infty}
\bigl(p_0Q_F^n-\pi Q_F^{n+1}\bigr)(A),\\
n_Q&\ge n_Q^*
:=
\sup_{s\in\mathbb R^d}
\frac{\max\{\pi(s),p_0(s)\}-u(s)}{\pi_Q(s)},
\end{aligned}
\end{equation}
and $u(s)$ is the density of the measure $U$.

2. The classifiers $d_F$, $d_B$ and the backward transition density $q_B$ satisfy
\begin{equation}
\begin{aligned}
d_F(s) &= \frac{r(s)}{f(s)}, \qquad d_B(s) = \frac{Z \, p_0(s)}{f(s)},\\
q_B(s \mid s') &= \frac{f(s)-r(s)}{f(s')-Z \, p_0(s')}q_F(s' \mid s),
\end{aligned}
\end{equation}
where $f(s) = Z(u(s) + n_Q \pi_Q(s))$ is the density of the flow measure $F$.
\end{theorem}
\begin{corollary}
\label{cor:conn_mcmc}
The constant $n_Q$ from \cref{thm:conn_mcmc} coincides with the mean trajectory length $\mathbb{E}_{\tau \sim P}[n_{\tau}]$. The total flow $F(\mathcal S)$ is minimized if and only if $n_Q = n^{*}_Q$.
\end{corollary}
We provide the proof of these statements in \cref{proof:conn_mcmc}. The measure \(U(A)\) admits an interpretation as the accumulated discrepancy between injected and removed mass from a measurable set \(A\), when we run a Markov chain with the fixed kernel \(Q_F\). Additional discussion and interpretation are provided in \cref{sec:conn_mcmc_interpret}. This result shows that, once the forward kernel \(Q_F\) is fixed, for example to the ULA kernel, we can potentially de-bias it to obtain samples from a target distribution \(\pi(x)\) using a classifier. However, doing so requires at least \(n^{*}_Q\) steps on average, given that this quantity is finite.

If the forward kernel \(Q_F\) is already invariant with respect to the target distribution, i.e., $\pi_Q = \pi$, then the measure $U$ simplifies to $U(A) = \sum_{n=0}^{\infty} (p_0 Q_F^n - \pi)(A)$,
which can be intuitively viewed as an accumulated discrepancy between distributions at step $n$ of the Markov chain $p_0 Q_F^n$ and the target $\pi$.

\subsection{Training algorithm}
Based on our theory, various GFlowNet objectives can be used to train the model. However, there are various considerations related to their practical efficiency. Detailed balance (DB)~\cite{bengio2023gflownet} is a local objective, and~\citet{morozov2025revisiting} showed that it can be effectively used with state flow regularization to train non-acyclic GFlowNets. However, we found DB to perform poorly in continuous environments, which is consistent with the findings of~\citet{berner2026from}. Alternatively, subtrajectory balance (SubTB)~\cite{madan2023learning} objective can be used, which incorporates both local and global trajectory information, but introduces a quadratic number of terms in the trajectory length, making it computationally inefficient for training in non-acyclic environments. Recently \citet{morozov2026learning} introduced a computationally efficient prefix trajectory balance objective, which we found well-suited for training in non-acyclic continuous spaces, and use it in our experiments.

We assume parametric forward and backward policies \(p_F\) and \(p_B\) with learnable classifiers \(d_F,d_B\) and learnable kernels \(q_F,q_B\). At each iteration, we sample partial trajectories of fixed length \(N_{\max}\) with the classifier disabled to encourage exploration. Each prefix of length $i$ is extended with a terminal transition, yielding a complete trajectory \(\tau_{0:i}\). For each prefix, we minimize
\begin{equation}
\label{eq:tb_loss}
\mathcal L_{\text{TB}}(\theta, \tau_{0:i})
=
\bigg(\log\frac{p_0(s_1)\prod_{t=1}^{i} p_F(s_{t+1}\mid s_t;\theta)}
{r(s_i)/Z_\theta \prod_{t=0}^{i-1} p_B(s_t\mid s_{t+1};\theta)}\bigg)^2,
\end{equation}
where the normalizing constant \(Z_\theta\) is treated as a learnable scalar parameter and allows the model to match the partition function $Z = R(\mathcal X)$.

To encourage shorter trajectories, we add flow regularization \citep{brunswic2024theory, morozov2025revisiting}, which can be computed directly from the classifier using \eqref{eq:flow_reward_clf}. 
We sum the loss over all prefixes, weighted by the model's stopping probability at step $i$: $w_i = \mathrm{sg}\!\left[ d_F(s_i;\theta)\prod_{t=1}^{i-1}(1-d_F(s_t;\theta))\right]$, where \(\mathrm{sg}\) denotes stopgrad. The weights focus training on prefixes that are likely under the current policy and thus improve stability. The final objective is
\begin{equation}
\mathcal L(\theta,\tau,\rho)
=
\sum_{i=1}^{N_{\max}} w_i
\left(
\mathcal L_{\mathrm{TB}}(\theta,\tau_{0:i})
+
\rho\,\frac{r(s_i)}{d_F(s_i;\theta)}
\right),
\end{equation}
where \(\rho>0\) controls flow regularization. 

In \cref{sec:param}, we provide the concrete parameterizations of the continuous kernels \(q_F\) and \(q_B\). Learning a parameterized backward policy is particularly natural in our setting, as it can help encourage shorter trajectories. This perspective is also supported by recent work of \citet{gritsaev2025adaptive}, who show that adaptive optimization of the backward policy can substantially improve diffusion sampler training.
\subsection{Multilevel generalization}
\label{sec:ml_algorithm}
We extend the state space with levels variables, allowing the model to switch between regimes during generation. In the basic formulation from \cref{sec:main_algorithm}, the transition kernels and classifiers are homogeneous and do not explicitly depend on the generation step. To introduce additional flexibility, we augment the state space and consider continuous states of the form \((s, \ell)\), where \(s \in \mathbb{R}^d\) and \(\ell \in \{1, \dots, L\}\), with \(L\) denoting the number of levels.

Starting from \(s_0\), we sample \(s\sim p_0\) at level \(\ell=1\) and evolve the process using \(P_F\). The classifier decides whether to continue within the current level or transition to the next one, until level \(L\) is reached and the trajectory terminates at \(s_f\). The backward dynamics are defined analogously. The resulting process remains non-acyclic, as cycles may occur within each level. Inspired by annealed importance sampling \citep[AIS;][]{neal2001annealed} and the use of geometric interpolations in dynamics-based samplers (e.g., \citet{mate2023learning,choi2026reinforced}), we define intermediate rewards
\begin{equation}
r(s,\ell)=p_0(s)^{1-\beta_\ell}r(s)^{\beta_\ell},
\end{equation}
where \(0=\beta_0<\cdots<\beta_L=1\). This yields a sequence of distributions interpolating between \(p_0\) and \(R\). For each pair \((\ell-1,\ell)\), where $l \in \{1,\dots,L\}$, we solve a GFlowNet problem with initial and target distributions
\begin{equation}
\begin{aligned}
p_0^{(\ell)}(s)&={r(s,\ell-1)}/{Z^{(\ell-1)}},\\
\pi^{(\ell)}(s)&={r(s,\ell)}/{Z^{(\ell)}}.
\end{aligned}
\end{equation}
These are implemented using shared forward and backward policies conditioned on the level \(\ell\), with learnable constants \(Z^{(\ell)}_{\theta}\). Training follows \cref{sec:main_algorithm}, where we sample trajectories of length \(N_{\max}\), allocate \(N_L=N_{\max}/L\) steps per level, and apply prefix trajectory balance within each segment. The final objective sums losses across all levels with the same weighting and flow regularization.

The multilevel scheme decomposes a difficult generation problem into a sequence of easier ones. Intermediate rewards \(r(s,\ell)\) are smoother and closer to each other than the original pair \((p_0, R)\), stabilizing learning and exploration. Correctness follows from the following evident statement.
\begin{proposition}
\label{prop:ml_tb_yields_tb}
If each pair of levels \((\ell-1,\ell)\), where $l \in \{1,\dots,L\}$, satisfies trajectory balance with initial distribution \(r(s,\ell-1)/Z^{(\ell-1)}\) and target \(r(s,\ell)/Z^{(\ell)}\), then the terminating distribution $p_T(A) = \pi(A)$.
\end{proposition}

\begin{remark}
A related line of work is diffusion-based sampling (e.g., \citet{zhang2021path,sendera2024improved}), where samples evolve through a sequence of intermediate distributions, moving to the next level after each transition. In contrast, our model can apply the same transition kernel multiple times per level, using a learned classifier to decide when to advance. This added flexibility allows the sampler to allocate more steps to difficult regions and fewer to simpler ones.
\end{remark}

\section{Experiments}
\paragraph{Improvement over ULA.}
\label{sec:exp_improve_ula}

\begin{figure}[b!]
\centering
\vspace*{-1em}
\includegraphics[width=0.9\linewidth]{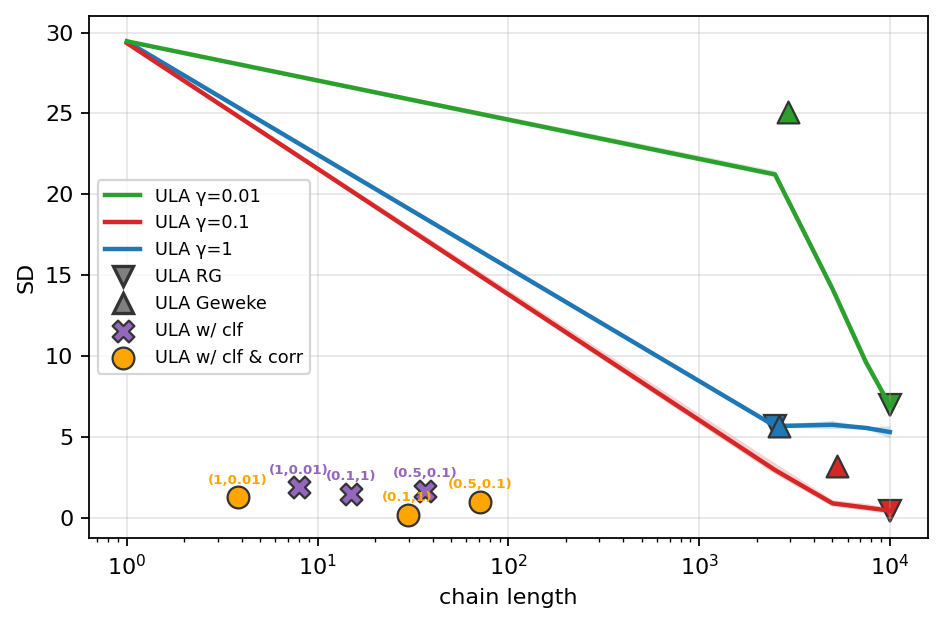}
\caption{Comparison of our approach to ULA and ULA with stopping based on the Gelman–Rubin and Geweke convergence diagnostics on GMM9 task, where modes are imbalanced. Pairs of two values next to points corresponding to our models denote $(\gamma, \rho)$, step size and regularization coefficient respectively. Baselines (ULA, ULA RG, ULA Geweke) with equal step sizes share the same color. \textit{ULA w/ clf} denotes our model with $q_F$ fixed to the ULA kernel. \textit{ULA w/ clf \& corr} denotes our full model with learnable $q_F$.}
\label{fig:improve_ula_steps_vs_metrics}
\end{figure}

\begin{figure*}[t]
\centering
\includegraphics[width=0.25\linewidth]{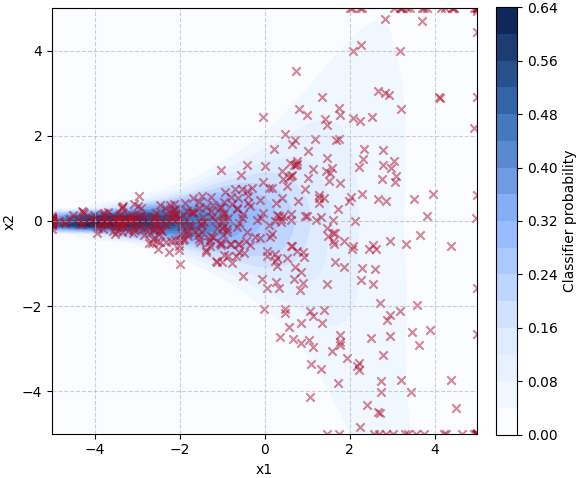}
\includegraphics[width=0.25\linewidth]{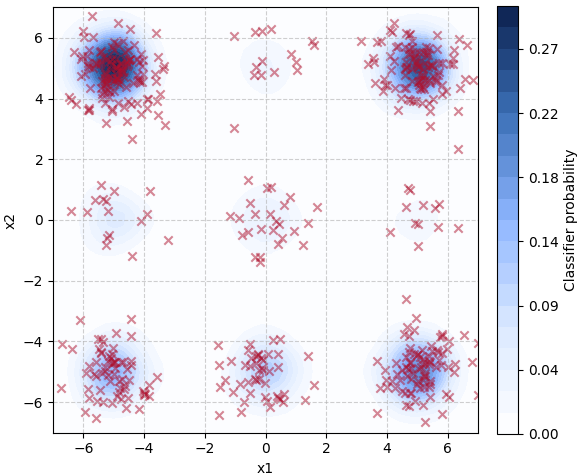}
\includegraphics[width=0.25\linewidth]{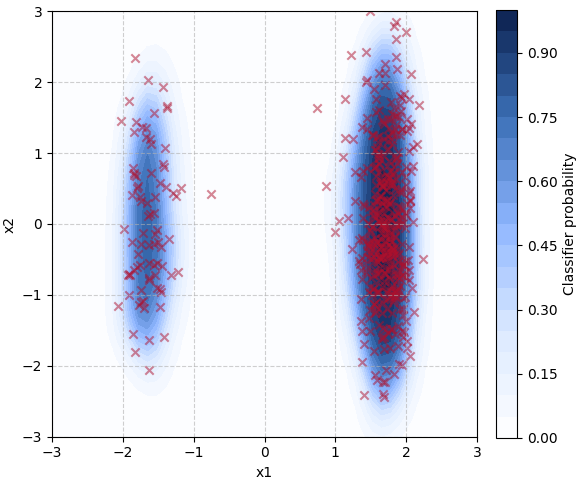}
\caption{
Forward classifier $d_F(s)$ visualizations for different environments (from left to right: Funnel, GMM9, ManyWell, with the first and last projected onto the first two coordinates).
}
\label{fig:improve_ula_fwd_classifier}
\vspace*{-0.5em}
\end{figure*}
\begin{figure*}[t]
\centering
\includegraphics[width=0.25\textwidth]{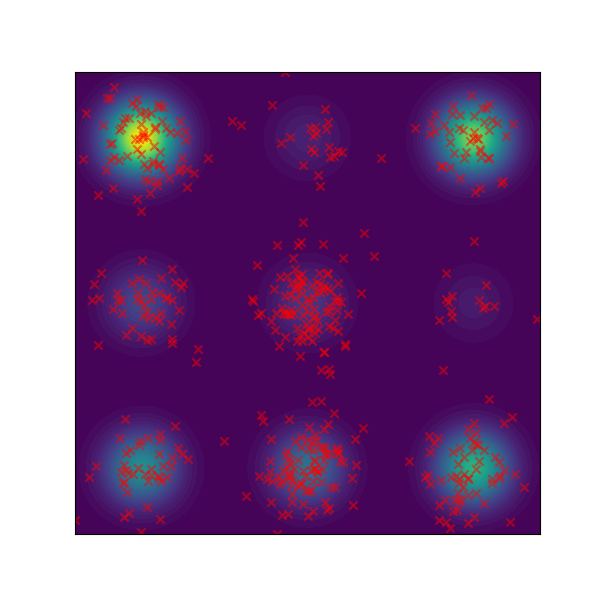}
\includegraphics[width=0.25\textwidth]{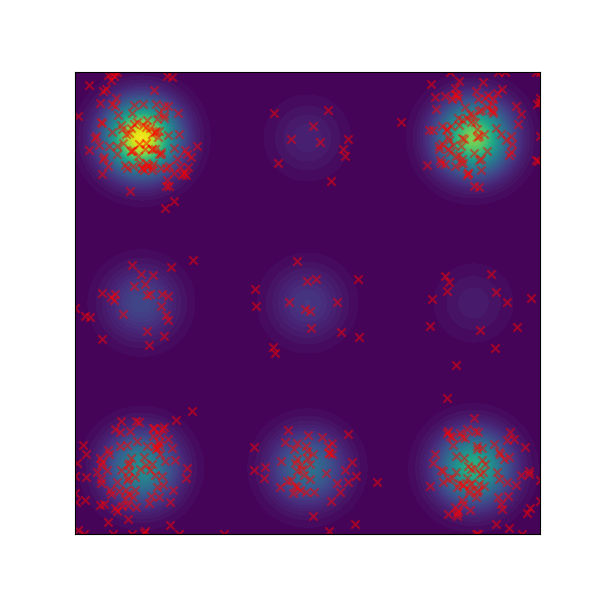}
\includegraphics[width=0.25\textwidth]{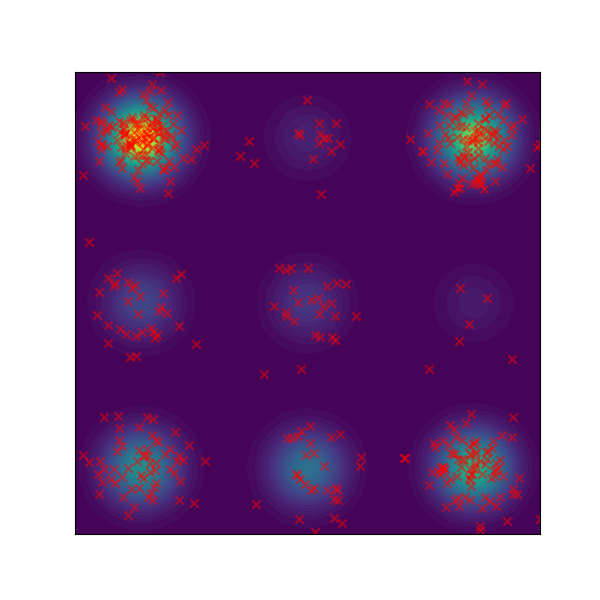}
\vspace*{-1em}
\caption{
Samples on the GMM9 target. Left to right: ULA, ULA with classifier, ULA with classifier and drift correction. True density shown in background. Additional images in \cref{sec:add_vis}.
}
\label{fig:samples_gmm9_2d}
\vspace*{-1em}
\end{figure*}

In this section, we evaluate the proposed method against the Unadjusted Langevin Algorithm (ULA) and study the effects of classifier-based early stopping and learned corrections to the forward dynamics. Experiments are performed on several synthetic environments described in \cref{sec:envs}. We evaluate the algorithms using standard metrics commonly adopted in the sampling literature, following \citet{blessing2024beyond}. Additional details are provided in \cref{sec:metrics}.

We first consider a configuration where the forward continuous kernel \(q_F\) is fixed to the ULA proposal, while the forward classifier \(d_F\), the backward classifier \(d_B\), and the backward kernel \(q_B\) are learned. In this setting, the learned classifier significantly reduces the average trajectory length relative to fixed-step ULA, as shown in \cref{tab:improve_ula_table_main}. However, the improvement in sample quality is generally limited. 

Next, we study the full model, where the forward kernel \(q_F\) is also learned (as a correction to the ULA kernel, see~\cref{sec:param}) in addition to \(d_F\), \(d_B\), and \(q_B\). In this setting, the sampler is able to make substantially more effective transitions between distant high-density regions, leading to improved approximation quality. The learned classifier in the full model is visualized in \cref{fig:improve_ula_fwd_classifier}. The classifier assigns higher stopping probabilities in regions of high density and lower probabilities elsewhere, effectively recovering the geometry of the target distribution. This behavior is also consistent with the relation \(d_F(s)=r(s)/f(s)\), since the flow typically becomes close to uniform within a mode, the classifier tends to resemble the target density itself.

\cref{fig:improve_ula_traj_compare} in the Appendix compares trajectories produced by ULA and the full model. As expected, ULA remains trapped within a single mode or local region and therefore requires a large time to move between modes. In contrast, the proposed method is able to make effective transitions between modes and more distant high-density regions due to learned stopping and corrections to the ULA dynamics.

\newcommand{\meanstd}[2]{$#1\scriptscriptstyle\pm#2$}

\begin{table*}[t]
\centering
\caption{\looseness=-1 Comparison of various ULA extensions across four target densities. The best result in each row is highlighted in gray. \textbf{ULA w/ clf} denotes our model with $q_F$ fixed to the ULA kernel. \textbf{ULA w/ clf \& corr} denotes our full model with learnable $q_F$. Mean$\pm$std over three runs.}
\label{tab:improve_ula_table_main}
\tiny
\setlength{\tabcolsep}{1.3pt}
%\resizebox{\linewidth}{!}{
\begin{tabular}{@{}l*{4}{cc}}
\toprule
Target $\rightarrow$
& \multicolumn{2}{c}{GMM9 ($d=2$)}
& \multicolumn{2}{c}{Funnel ($d=2$)}
& \multicolumn{2}{c}{GMM9 ($d=40$)}
& \multicolumn{2}{c}{ManyWell ($d=40$)} \\
\cmidrule(lr){2-3}\cmidrule(lr){4-5}\cmidrule(lr){6-7}\cmidrule(lr){8-9}
Algorithm $\downarrow$ Metric $\rightarrow$
& SD $\downarrow$ & MMD $\downarrow$
& SD $\downarrow$ & MMD $\downarrow$
& SD $\downarrow$ & MMD $\downarrow$
& SD $\downarrow$ & MMD $\downarrow$ \\
\midrule
ULA
& \meanstd{0.463}{0.131} & \meanstd{0.002}{0.005}
& \meanstd{14.101}{0.243} & \meanstd{0.142}{0.014}
& \meanstd{376.406}{2.165} & \meanstd{0.347}{0.011}
& \meanstd{71.508}{0.876} & \meanstd{0.332}{0.018} \\
ULA RG
& \meanstd{0.463}{0.154} & \cellcolor{bestcolor} \meanstd{0.002}{0.001}
& \meanstd{11.686}{0.165} & \meanstd{0.112}{0.002}
& \meanstd{376.652}{2.377} & \meanstd{0.345}{0.002}
& \meanstd{70.614}{0.189} & \meanstd{0.328}{0.001} \\
ULA Geweke
& \meanstd{3.234}{0.243} & \meanstd{0.010}{0.001}
& \meanstd{16.262}{0.133} & \meanstd{0.216}{0.004}
& \meanstd{376.407}{1.935} & \meanstd{0.346}{0.001}
& \meanstd{73.531}{0.081} & \meanstd{0.338}{0.001} \\
\midrule 
\textbf{ULA w/ clf}
& \meanstd{1.522}{0.051} & \meanstd{0.053}{0.002}
& \meanstd{8.849}{0.126} & \meanstd{0.128}{0.003}
& \meanstd{321.803}{2.038} & \meanstd{0.314}{0.017}
& \cellcolor{bestcolor} \meanstd{54.804}{4.203} & \cellcolor{bestcolor} \meanstd{0.298}{0.019} \\
\textbf{ULA w/ clf \& corr}
& \cellcolor{bestcolor}\meanstd{0.221}{0.026} & \meanstd{0.027}{0.006}
& \cellcolor{bestcolor}\meanstd{4.947}{0.097} & \cellcolor{bestcolor}\meanstd{0.089}{0.013}
& \cellcolor{bestcolor}\meanstd{243.809}{1.984} & \cellcolor{bestcolor}\meanstd{0.251}{0.019}
& \meanstd{62.472}{1.237} & \meanstd{0.314}{0.017} \\
\midrule
Algorithm $\downarrow$ Metric $\rightarrow$
& ELBO $\uparrow$ & EUBO $\downarrow$
& ELBO $\uparrow$ & EUBO $\downarrow$
& ELBO $\uparrow$ & EUBO $\downarrow$
& ELBO $\uparrow$ & EUBO $\downarrow$ \\
\midrule
\textbf{ULA w/ clf}
& \meanstd{-1.753}{0.045} & \meanstd{2.478}{0.083}
& \meanstd{-1.264}{0.038} & \meanstd{14.462}{0.312}
& \meanstd{-64.581}{0.912} & \meanstd{174.390}{1.726}
& \meanstd{92.507}{10.109} & \meanstd{317.601}{15.301} \\
\textbf{ULA w/ clf \& corr}
& \cellcolor{bestcolor}\meanstd{-0.157}{0.025} & \cellcolor{bestcolor}\meanstd{2.278}{0.065}
& \cellcolor{bestcolor}\meanstd{-0.392}{0.023} & \cellcolor{bestcolor}\meanstd{1.005}{0.053}
& \cellcolor{bestcolor}\meanstd{-51.701}{0.778} & \cellcolor{bestcolor}\meanstd{168.609}{1.618}
& \cellcolor{bestcolor}\meanstd{150.008}{8.421} & \cellcolor{bestcolor}\meanstd{248.303}{9.152} \\
\midrule
Algorithm $\downarrow$ Metric $\rightarrow$
& \multicolumn{2}{c}{\(\mathbb{E}[{n_{\tau}}]\downarrow\)}
& \multicolumn{2}{c}{\(\mathbb{E}[{n_{\tau}}]\downarrow\)}
& \multicolumn{2}{c}{\(\mathbb{E}[{n_{\tau}}]\downarrow\)}
& \multicolumn{2}{c}{\(\mathbb{E}[{n_{\tau}}]\downarrow\)} \\
\midrule
ULA
& \multicolumn{2}{c}{$10^4$}
& \multicolumn{2}{c}{$5 \times 10^2$}
& \multicolumn{2}{c}{$5 \times 10^2$}
& \multicolumn{2}{c}{$1.5 \times 10^3$} \\
ULA RG
& \multicolumn{2}{c}{$10^4$}
& \multicolumn{2}{c}{$10^3$}
& \multicolumn{2}{c}{$10^3$}
& \multicolumn{2}{c}{$2 \times 10^3$} \\
ULA Geweke
& \multicolumn{2}{c}{\meanstd{5307.304}{29.689}}
& \multicolumn{2}{c}{\meanstd{406.579}{0.662}}
& \multicolumn{2}{c}{\meanstd{437.228}{1.370}} 
& \multicolumn{2}{c}{\meanstd{1468.161}{8.509}} \\
\midrule
\textbf{ULA w/ clf}
& \multicolumn{2}{c}{\cellcolor{bestcolor}\meanstd{15.017}{0.227}}
& \multicolumn{2}{c}{\cellcolor{bestcolor}\meanstd{21.842}{0.330}}
& \multicolumn{2}{c}{\meanstd{37.659}{2.589}}
& \multicolumn{2}{c}{\meanstd{79.123}{4.203}} \\
\textbf{ULA w/ clf \& corr}
& \multicolumn{2}{c}{\meanstd{29.591}{0.345}}
& \multicolumn{2}{c}{\meanstd{36.337}{0.421}}
& \multicolumn{2}{c}{\cellcolor{bestcolor}\meanstd{27.837}{3.494}}
& \multicolumn{2}{c}{\cellcolor{bestcolor}\meanstd{75.383}{2.918}} \\
\bottomrule
\end{tabular}
%}
\end{table*}

\begin{table*}[t]
\centering
\caption{Comparison of the multilevel GFlowNet with levels $L=\{1,2,5\}$ against diffusion sampling baselines PIS and DDS on GMM9 (\(d=40\)), ManyWell (\(d=40\)), and Digits (\(d=196\)).}
\label{tab:ml_table_main}
\tiny
\setlength{\tabcolsep}{5pt}
\begin{tabular}{@{}l*{3}{cc}}
\toprule
Target $\rightarrow$
& \multicolumn{2}{c}{GMM9 ($d=40$)} 
& \multicolumn{2}{c}{ManyWell ($d=40$)} 
& \multicolumn{2}{c}{Digits ($d=196$)}\\
\cmidrule(lr){2-3}\cmidrule(lr){4-5}\cmidrule(lr){6-7}
Algorithm $\downarrow$ Metric $\rightarrow$
& SD $\downarrow$ & MMD $\downarrow$
& SD $\downarrow$ & MMD $\downarrow$
& SD $\downarrow$ & MMD $\downarrow$\\
\midrule
PIS
& \meanstd{329.400}{1.814} & \meanstd{0.186}{0.012}
& \meanstd{38.740}{1.329} & \meanstd{0.258}{0.011}
& \meanstd{1728.000}{10.351} & \meanstd{0.540}{0.023}
\\
DDS
& \meanstd{180.200}{1.310} & \cellcolor{bestcolor} \meanstd{0.181}{0.015}
& \meanstd{38.81}{1.329} & \meanstd{0.259}{0.013}
& \cellcolor{bestcolor} \meanstd{202.400}{4.189} & \meanstd{0.346}{0.021}
\\
\midrule 
\textbf{ULA w/ clf \& corr}
& \meanstd{243.809}{1.984} & \meanstd{0.251}{0.019}
& \meanstd{62.472}{1.237} & \meanstd{0.314}{0.017}
& \meanstd{469.500}{1.532} & \meanstd{0.410}{0.036}
\\
\textbf{Multilevel ($L=2$)}
& \meanstd{186.200}{1.723} & \meanstd{0.203}{0.015}
& \meanstd{36.330}{0.561} & \meanstd{0.108}{0.021}
& \meanstd{343.600}{0.386} & \meanstd{0.394}{0.042}
\\
\textbf{Multilevel ($L=5$)}
& \cellcolor{bestcolor} \meanstd{152.347}{2.281} & \cellcolor{bestcolor} \meanstd{0.184}{0.011}
& \cellcolor{bestcolor} \meanstd{28.400}{4.843} & \cellcolor{bestcolor} \meanstd{0.095}{0.020}
& \cellcolor{bestcolor} \meanstd{196.938}{3.519} & \cellcolor{bestcolor} \meanstd{0.304}{0.028}\\
\midrule
Algorithm $\downarrow$ Metric $\rightarrow$
& ELBO $\uparrow$ & EUBO $\downarrow$
& ELBO $\uparrow$ & EUBO $\downarrow$
& ELBO $\uparrow$ & EUBO $\downarrow$\\
\midrule
PIS
& \meanstd{-278.4}{1.841} & \meanstd{47.55}{4.125}
& \meanstd{200.9}{7.924} & \meanstd{278.9}{5.118}
& \meanstd{-3671.000}{21.931} & \meanstd{925.200}{7.815}
\\
DDS
& \cellcolor{bestcolor} \meanstd{-7.631}{5.183} 
& \cellcolor{bestcolor} \meanstd{6.610}{9.131}
& \cellcolor{bestcolor} \meanstd{201.100}{8.914} 
& \meanstd{331.100}{7.185}
& \meanstd{-149.300}{3.194} 
& \meanstd{149.9}{8.917} \\
\midrule
\textbf{ULA w/ clf \& corr}
& \meanstd{-51.701}{0.778} & \meanstd{168.609}{1.618}
& \meanstd{150.008}{8.421} & \meanstd{248.303}{9.152}
& \meanstd{-210.340}{3.591} & \meanstd{270.913}{5.951} \\
\textbf{Multilevel ($L=2$)}
& \meanstd{-34.735}{6.192} & \meanstd{71.100}{6.813}
& \meanstd{143.500}{4.254} & \meanstd{254.9}{3.138}
& \meanstd{-184.391}{8.138} & \meanstd{138.200}{6.984} \\
\textbf{Multilevel ($L=5$)}
& \meanstd{-28.421}{2.156} & \meanstd{58.734}{3.021}
& \meanstd{154.267}{3.845} & \cellcolor{bestcolor} \meanstd{221.456}{4.932}
& \cellcolor{bestcolor} \meanstd{-121.570}{4.193} & \cellcolor{bestcolor} \meanstd{102.630}{3.903} \\
\midrule
Algorithm $\downarrow$ Metric $\rightarrow$
& \multicolumn{2}{c}{\(\mathbb{E}[{n_{\tau}}]\downarrow\)}
& \multicolumn{2}{c}{\(\mathbb{E}[{n_{\tau}}]\downarrow\)}
& \multicolumn{2}{c}{\(\mathbb{E}[{n_{\tau}}]\downarrow\)}\\
\midrule
PIS
& \multicolumn{2}{c}{45}
& \multicolumn{2}{c}{75}
& \multicolumn{2}{c}{65}
\\
DDS
& \multicolumn{2}{c}{45}
& \multicolumn{2}{c}{75}
& \multicolumn{2}{c}{65} \\
\midrule
\textbf{ULA w/ clf \& corr}
& \multicolumn{2}{c}{\cellcolor{bestcolor} \meanstd{27.837}{3.494}}
& \multicolumn{2}{c}{\meanstd{75.383}{2.918}} 
& \multicolumn{2}{c}{\meanstd{71.600}{6.139}} \\
\textbf{Multilevel ($L=2$)}
& \multicolumn{2}{c}{\meanstd{44.600}{3.613}}
& \multicolumn{2}{c}{\meanstd{74.173}{2.590}}
& \multicolumn{2}{c}{\cellcolor{bestcolor} \meanstd{62.300}{5.058}} \\
\textbf{Multilevel ($L=5$)}
& \multicolumn{2}{c}{\meanstd{72.900}{4.152}}
& \multicolumn{2}{c}{\cellcolor{bestcolor} \meanstd{68.730}{2.741}}
& \multicolumn{2}{c}{\meanstd{68.281}{5.300}} \\
\bottomrule
\end{tabular}
\vspace{-0.3cm}
\end{table*}

The quantitative results are summarized in \cref{tab:improve_ula_table_main}. Evaluation metrics are Sinkhorn distance (SD), RBF maximum mean discrepancy (MMD), evidence lower and upper bounds (ELBO, EUBO), and the mean trajectory length $\mathbb{E}[n_{\tau}]$. ULA RG and ULA Geweke denote ULA with stopping based on the Gelman--Rubin and Geweke convergence diagnostics, respectively. Overall, classifier-based stopping substantially improves sampling efficiency by reducing the average trajectory length, while learned forward corrections further improve approximation quality, typically achieving lower Sinkhorn distance and MMD, and tighter ELBO/EUBO estimates. For ULA, we report metrics after the chain reaches a stationary regime with respect to the Sinkhorn metric. In~\cref{fig:improve_ula_steps_vs_metrics}, we compare our methods against ULA and standard stopping diagnostics across different step sizes and numbers of steps on a gaussian mixture target where modes have different weights. Even on the relatively simple two-dimensional imbalanced mixture task, MCMC-based approaches remain impractical, as they require extremely long mixing times to correctly recover the target distribution, while the learned classifier helps to correctly adjust to the weights of different mixture components. This behavior is also clearly seen in the visualization of the samples in Figure~\ref{fig:samples_gmm9_2d}. Additional experiments, hyperparameters and visualizations are provided in \cref{sec:appendix_exp_improve_ula}.

\paragraph{Experiments with the multilevel scheme.}
\label{sec:exp_ml}
We now evaluate the proposed multilevel extension described in \cref{sec:ml_algorithm}. We consider configurations with $L \in \{1,2,5\}$ levels and compare them against established diffusion sampling methods, namely PIS~\cite{zhang2021path} and DDS~\cite{vargas2023denoising}. Note that a multilevel scheme with $L=1$ reduces to the single-level scheme with learnable $q_F$.

\cref{tab:ml_table_main} reports results on the GMM9 and ManyWell targets with $d=40$, as well as on the more challenging $d=196$ dimensional Digits task from~\citet{blessing2024beyond}, corresponding to sampling MNIST images from density induced by a NICE normalizing flow model~\cite{dinh2014nice}. For a trained multilevel sampler, we measure it mean trajectory length and train diffusion samplers with the approximately this number of steps. Overall, the multilevel scheme consistently improves sample quality relative to the single-level model, especially on more difficult and high-dimensional tasks. Increasing the number of levels further improves approximation quality, yielding lower SD, MMD, and tighter ELBO/EUBO estimates. On GMM9 and Digits, the multilevel models significantly outperform the single-level baseline and become competitive with diffusion samplers. In particular, the \(L=5\) configuration achieves the best overall performance among the proposed methods.

\section{Conclusion}
\label{sec:conclusion}

We have proposed a novel framework for adaptive sampling that treats trajectory termination as a learnable component of the dynamics. By casting MCMC with stopping decisions into the framework of non-acyclic GFlowNets, we established a principled connection between optimal stopping policies and the target distribution via flow and detailed balance conditions. Empirically, the method significantly reduces trajectory lengths while improving mode coverage and mixing across a range of standard benchmarks. The proposed approach introduces additional learning components, which increases optimization complexity and can require careful balancing between exploration and regularization during training, posing typical challenges that arise in learning-based sampling methods. Future work could focus on improving training objectives for continuous non-acyclic GFlowNets, as well as extending the framework to more expressive sampling dynamics, e.g., Hamiltonian Monte Carlo~\cite{neal2011mcmc}.

\section*{Acknowledgments}
This research was supported in part through computational resources of HPC facilities at HSE University~\citep{kostenetskiy2021hpc}. ESW acknowledges support from the CIFAR Learning in Machines and Brains programme.

\section*{Impact Statement}
This paper presents work whose goal is to advance the field of Machine
Learning. There are many potential societal consequences of our work, none
which we feel must be specifically highlighted here.

\bibliography{refs}

@article{neal2001annealed,
  title={Annealed importance sampling},
  author={Neal, Radford M},
  journal={Statistics and computing},
  volume={11},
  number={2},
  pages={125--139},
  year={2001},
  publisher={Springer}
}

@inproceedings{kostenetskiy2021hpc,
  title={HPC resources of the higher school of economics},
  author={Kostenetskiy, PS and Chulkevich, RA and Kozyrev, VI},
  booktitle={Journal of Physics: Conference Series},
  volume={1740},
  number={1},
  pages={012050},
  year={2021},
  organization={IOP Publishing}
}

@book{douc:moulines:priouret:soulier:2018,
    AUTHOR = {Douc, R. and Moulines, E. and Priouret, P. and Soulier, P.},
     TITLE = {Markov chains},
    SERIES = {Springer Series in Operations Research and Financial Engineering},
 PUBLISHER = {Springer},
      YEAR = {2018}
}

@article{dinh2014nice,
  title={{NICE}: Non-linear Independent Components Estimation},
  author={Dinh, Laurent and Krueger, David and Bengio, Yoshua},
  journal={arXiv preprint arXiv:1410.8516},
  year={2014}
}

@article{hoffman2014no,
  title={The {No-U-Turn} sampler: adaptively setting path lengths in {Hamiltonian Monte Carlo}},
  author={Hoffman, Matthew D and Gelman, Andrew and others},
  journal={Journal of Machine Learning Research},
  volume={15},
  number={1},
  pages={1593--1623},
  year={2014}
}

@article{neal2011mcmc,
  title={{MCMC} using {Hamiltonian} dynamics},
  author={Neal, Radford M and others},
  journal={Handbook of Markov Chain Monte Carlo},
  volume={2},
  number={11},
  pages={2},
  year={2011},
  publisher={Chapman and Hall/CRC}
}

@article{gabrie2022adaptive,
  title={Adaptive {Monte Carlo} augmented with normalizing flows},
  author={Gabri{\'e}, Marylou and Rotskoff, Grant M and Vanden-Eijnden, Eric},
  journal={Proceedings of the National Academy of Sciences},
  volume={119},
  number={10},
  pages={e2109420119},
  year={2022},
  publisher={National Academy of Sciences}
}

@article{andrieu2003introduction,
  title={An introduction to {MCMC} for machine learning},
  author={Andrieu, Christophe and De Freitas, Nando and Doucet, Arnaud and Jordan, Michael I},
  journal={Machine learning},
  volume={50},
  number={1},
  pages={5--43},
  year={2003},
  publisher={Springer}
}

@book{liu2001monte,
  title={{Monte Carlo} strategies in scientific computing},
  author={Liu, Jun S and Liu, Jun S},
  volume={10},
  year={2001},
  publisher={Springer}
}

@article{welling2011bayesian,
  title={Bayesian learning via stochastic gradient {Langevin} dynamics},
  author={Welling, Max and Teh, Yee W},
  booktitle={International Conference on Machine Learning (ICML)},
  year={2011}
}

@article{noe2019boltzmann,
  title={Boltzmann generators: Sampling equilibrium states of many-body systems with deep learning},
  author={No{\'e}, Frank and Olsson, Simon and K{\"o}hler, Jonas and Wu, Hao},
  journal={Science},
  volume={365},
  number={6457},
  year={2019},
  publisher={American Association for the Advancement of Science}
}

@article{madan2023learning,
  title={Learning {GFlowNets} from partial episodes for improved convergence and stability},
  author={Madan, Kanika and Rector-Brooks, Jarrid and Korablyov, Maksym and Bengio, Emmanuel and Jain, Moksh and Nica, Andrei Cristian and Bosc, Tom and Bengio, Yoshua and Malkin, Nikolay},
  journal={International Conference on Machine Learning (ICML)},
  year={2023}
}

@article{izmailov2021bayesian,
  title={What are {Bayesian} neural network posteriors really like?},
  author={Izmailov, Pavel and Vikram, Sharad and Hoffman, Matthew D and Wilson, Andrew Gordon Gordon},
  journal={International Conference on Machine Learning (ICML)},
  year={2021},
}

@book{stoltz2010free,
  title={Free energy computations: A mathematical perspective},
  author={Stoltz, Gabriel and Rousset, Mathias and others},
  year={2010},
  publisher={World Scientific}
}

@article{gretton2012kernel,
  title={A kernel two-sample test},
  author={Gretton, Arthur and Borgwardt, Karsten M and Rasch, Malte J and Sch{\"o}lkopf, Bernhard and Smola, Alexander},
  journal={Journal of Machine Learning Research},
  volume={13},
  number={1},
  pages={723--773},
  year={2012}
}

@book{peyre2019computational,
  title={Computational optimal transport: With applications to data science},
  author={Peyr{\'e}, Gabriel and Cuturi, Marco},
  year={2019},
  publisher={Now Foundations and Trends}
}

@article{
zhang2021path,
title={Path Integral Sampler: A Stochastic Control Approach For Sampling},
author={Zhang, Qinsheng and Chen, Yongxin},
journal={International Conference on Learning Representations (ICLR)},
year={2022}
}

@article{bengio2021flow,
  title={Flow network based generative models for non-iterative diverse candidate generation},
  author={Bengio, Emmanuel and Jain, Moksh and Korablyov, Maksym and Precup, Doina and Bengio, Yoshua},
  journal={Neural Information Processing Systems (NeurIPS)},
  year={2021}
}

@article{tiapkin2024generative,
  title={Generative flow networks as entropy-regularized {RL}},
  author={Tiapkin, Daniil and Morozov, Nikita and Naumov, Alexey and Vetrov, Dmitry P},
  journal={Artificial Intelligence and Statistics (AISTATS)},
  year={2024},
}

@article{blessing2024beyond,
  title={Beyond {ELBOs}: A Large-Scale Evaluation of Variational Methods for Sampling},
  author={Blessing, Denis and Jia, Xiaogang and Esslinger, Johannes and Vargas, Francisco and Neumann, Gerhard},
  journal={International Conference on Machine Learning (ICML)},
  year={2024},
}

@article{
vargas2023denoising,
title={Denoising Diffusion Samplers},
author={Vargas, Francisco and Grathwohl, Will and Doucet, Arnaud},
journal={International Conference on Learning Representations (ICLR)},
year={2023}
}

@article{
midgley2023flow,
title={Flow Annealed Importance Sampling Bootstrap},
author={Laurence Illing Midgley and Vincent Stimper and Gregor N. C. Simm and Bernhard Sch{\"o}lkopf and Jos{\'e} Miguel Hern{\'a}ndez-Lobato},
journal={International Conference on Learning Representations (ICLR)},
year={2023}
}

@article{bengio2023gflownet,
  title={{GFlowNet} foundations},
  author={Bengio, Yoshua and Lahlou, Salem and Deleu, Tristan and Hu, Edward J and Tiwari, Mo and Bengio, Emmanuel},
  journal={Journal of Machine Learning Research},
  volume={24},
  number={210},
  pages={1--55},
  year={2023}
}

@article{lahlou2023theory,
  title={A theory of continuous generative flow networks},
  author={Lahlou, Salem and Deleu, Tristan and Lemos, Pablo and Zhang, Dinghuai and Volokhova, Alexandra and Hern{\'a}ndez-Garc{\i}a, Alex and Ezzine, L{\'e}na N{\'e}hale and Bengio, Yoshua and Malkin, Nikolay},
  journal={International Conference on Machine Learning (ICML)},
  year={2023},
}

@article{zhang2023diffusion,
  title={Diffusion generative flow samplers: Improving learning signals through partial trajectory optimization},
  author={Zhang, Dinghuai and Chen, Ricky TQ and Liu, Cheng-Hao and Courville, Aaron and Bengio, Yoshua},
  journal={International Conference on Learning Representations (ICLR)},
  year={2024}
}

@article{sendera2024improved,
  title={Improved off-policy training of diffusion samplers},
  author={Sendera, Marcin and Kim, Minsu and Mittal, Sarthak and Lemos, Pablo and Scimeca, Luca and Rector-Brooks, Jarrid and Adam, Alexandre and Bengio, Yoshua and Malkin, Nikolay},
  journal={Neural Information Processing Systems (NeurIPS)},
  year={2024}
}

@article{brunswic2024theory,
  title={A theory of non-acyclic generative flow networks},
  author={Brunswic, Leo and Li, Yinchuan and Xu, Yushun and Feng, Yijun and Jui, Shangling and Ma, Lizhuang},
  journal={Association for the Advancement of Artificial Intelligence (AAAI)},
  year={2024}
}

@article{morozov2025revisiting,
  title={Revisiting Non-Acyclic {GFlowNets} in Discrete Environments},
  author={Morozov, Nikita and Maksimov, Ian and Tiapkin, Daniil and Samsonov, Sergey},
  journal={International Conference on Machine Learning (ICML)},
  year={2025},
}

@article{gritsaev2025adaptive,
  title={Adaptive destruction processes for diffusion samplers},
  author={Gritsaev, Timofei and Morozov, Nikita and Tamogashev, Kirill and Tiapkin, Daniil and Samsonov, Sergey and Naumov, Alexey and Vetrov, Dmitry and Malkin, Nikolay},
  journal={arXiv preprint arXiv:2506.01541},
  year={2025}
}

@article{morozov2026learning,
  title={Learning Shortest Paths with Generative Flow Networks},
  author={Morozov, Nikita and Maksimov, Ian and Tiapkin, Daniil and Samsonov, Sergey},
  journal={arXiv preprint arXiv:2603.01786},
  year={2026}
}

@article{mcmcdiag2019,
	author = {Roy, Vivekananda},
	journal = {Annual Review of Statistics and Its Application},
	pages = {387-412},
	title = {Convergence Diagnostics for {Markov} Chain {Monte Carlo}},
	volume = {7},
	year = {2020}
}

@misc{mcmcstopping2023,
  author = {Kwon, Sunbeom and Zhang, Susu and Köhn, Hans Friedrich and Zhang, Bo},
title = {{MCMC} stopping rules in latent variable modelling},
journal = {British Journal of Mathematical and Statistical Psychology},
volume = {78},
number = {1},
pages = {225-257},
year={2025}
}

@article{mcstopreview2025,
  author = {Jiezhong Wu and Reiichiro Kawai},
  title = {Stopping Rules for {Monte Carlo} Methods: A Review},
  year = {2025},
  journal={arXiv preprint arXiv:2510.22688}
}

@article{gradadaptmcmc2019,
  author = {Michalis Titsias and Petros Dellaportas},
  title = {Gradient-based Adaptive {Markov} Chain {Monte Carlo}},
  year = {2019},
  journal = {Neural Information Processing Systems (NeurIPS)}
}

@article{posdepmala2023,
  author = {Vivekananda Roy and Lijin Zhang},
  title = {Convergence of position-dependent {MALA} with application to conditional simulation in {GLMMs}},
  journal = {Journal of Computational and Graphical Statistics},
volume = {32},
number = {2},
pages = {501--512},
year = {2023},
}

@article{automala2023,
  author = {Miguel Biron-Lattes and Nikola Surjanovic and Saifuddin Syed and Trevor Campbell and Alexandre Bouchard-Cote},
  title = {{autoMALA}: Locally adaptive {Metropolis}-adjusted {Langevin} algorithm},
  year = {2023},
  journal={Artificial Intelligence and Statistics (AISTATS)}
}

@article{autostep2024,
  author = {Tiange Liu and Nikola Surjanovic and Miguel Biron-Lattes and Alexandre Bouchard-Cote and Trevor Campbell},
  title = {{AutoStep}: Locally adaptive involutive {MCMC}},
  year = {2025},
  journal = {International Conference on Machine Learning (ICML)}
}

@article{wang2024rlmh,
  author = {Congye Wang and Wilson Chen and Heishiro Kanagawa and Chris J. Oates},
  title = {Reinforcement Learning for Adaptive {MCMC}},
  year = {2025},
  journal={Artificial Intelligence and Statistics (AISTATS)}
}

@article{wang2025harnessing,
  author = {Congye Wang and Matthew A. Fisher and Heishiro Kanagawa and Wilson Chen and Chris J. Oates},
  title = {Harnessing the Power of Reinforcement Learning for Adaptive {MCMC}},
  year = {2025},
  journal = {arXiv preprint arXiv:2507.00671}
}

@article{
berner2026from,
title={From discrete-time policies to continuous-time diffusion samplers: Asymptotic equivalences and faster training},
author={Julius Berner and Lorenz Richter and Marcin Sendera and Jarrid Rector-Brooks and Nikolay Malkin},
journal={Transactions on Machine Learning Research},
year={2026},
}

@article{kim2025scalable,
      title={On scalable and efficient training of diffusion samplers}, 
      author={Minkyu Kim and Kiyoung Seong and Dongyeop Woo and Sungsoo Ahn and Minsu Kim},
      journal={Neural Information Processing Systems (NeurIPS)},
      year={2025}
}

@article{deleu2023generative,
      title={Generative Flow Networks: a Markov Chain Perspective}, 
      author={Tristan Deleu and Yoshua Bengio},
      year={2023},
      journal={arXiv preprint arXiv:2307.01422}
}

@article{choi2026reinforced,
  title={Reinforced sequential {M}onte {C}arlo for amortised sampling},
  author={Choi, Sanghyeok and Mittal, Sarthak and Elvira, V{\'i}ctor and Park, Jinkyoo and Whitammer, Esmeralda S},
  journal={International Conference on Machine Learning (ICML)},
  year={2026}
}

@article{mate2023learning,
    title={Learning Interpolations between {Boltzmann} Densities},
    author={B{\'a}lint M{\'a}t{\'e} and Fran{\c{c}}ois Fleuret},
    journal={Transactions on Machine Learning Research (TMLR)},
    year={2023},
}

@article{andrieu2008adaptive,
	author = {Andrieu, Christophe and Thoms, Johannes},
	journal = {Statistics and Computing},
	number = {4},
	pages = {343--373},
	title = {A tutorial on adaptive {MCMC}},
	volume = {18},
	year = {2008}
}

@article{vihola2012robust,
	author = {Vihola, Matti},
	journal = {Statistics and Computing},
	number = {5},
	pages = {997--1008},
	title = {Robust adaptive Metropolis algorithm with coerced acceptance rate},
	volume = {22},
	year = {2012}
}

@article{haario2001adaptive,
author = {Heikki Haario and Eero Saksman and Johanna Tamminen},
title = {{An adaptive Metropolis algorithm}},
volume = {7},
journal = {Bernoulli},
number = {2},
publisher = {Bernoulli Society for Mathematical Statistics and Probability},
pages = {223 -- 242},
year = {2001},
}

@article{vaswani2017attention,
  title={Attention is all you need},
  author={Vaswani, Ashish and Shazeer, Noam and Parmar, Niki and Uszkoreit, Jakob and Jones, Llion and Gomez, Aidan N and Kaiser, {\L}ukasz and Polosukhin, Illia},
  journal={Advances in neural information processing systems},
  volume={30},
  year={2017}
}
\bibliographystyle{icml2026}

\appendix
\onecolumn
\section{Experimental details}
\label{app:exp_details}
\subsection{Continuous kernels parameterizations}
\label{sec:param}
We initialize trajectories from the Gaussian distribution
\[
p_0(s)=\mathcal N(s\mid 0,\sigma_0^2 I).
\]
For transitions between continuous states, we use the forward density
\begin{equation}
\begin{aligned}
q_F(s' \mid s; \theta)&=\mathcal N\!\left(
s' \,\middle|\,
\mu_F(s;\theta),
\, \Sigma_F(s;\theta)
\right),\\
\mu_F(s;\theta) &= s + \gamma \Bigl(u_F(s;\theta) + \bigl(\mathbf 1+v_F(s;\theta)\bigr) \odot \nabla \log r(s)\Bigr),\\
\Sigma_F(s;\theta) &= 2 \gamma \,\mathrm{diag}\!\bigl(\exp \bigl \{C_F\tanh(w_F(s;\theta))\bigr \}\bigr),
\end{aligned}
\end{equation}
where \(u_F,v_F,w_F:\mathbb R^d\to\mathbb R^d\) are learnable functions, $\mathbf 1$ is a vector of $d$ ones, $\odot$ is a Hadamard product, \(\gamma\) is the step size and we set $C_F=4$ across all experiments.
The forward transition density \(q_F\) is chosen following the Langevin-type parameterization used in prior works \citep{zhang2021path}, \citep{sendera2024improved}. In particular, it resembles a discretization of Langevin dynamics, augmented with a learned correction to the drift term and a state-dependent scaling of the target score \(\nabla \log r(s)\). We also found it beneficial to learn the variance of the forward process, which provides additional adaptability compared to fixed-noise parameterizations. Note that when \(u_F \equiv 0\), \(v_F \equiv 0\), and \(w_F \equiv 0\), the forward transition reduces to the standard Unadjusted Langevin Algorithm (ULA) kernel.

For the backward transitions, we use the density
\begin{equation}
\begin{aligned}
q_B(s \mid s'; \theta)&=
\mathcal N\!\left(
s \,\middle|\,
\mu_B(s';\theta),
\, \Sigma_B(s';\theta)
\right),\\
\mu_B(s';\theta) &= s' - \gamma \,\mathrm{softplus}(u_B(s';\theta))\,s',\\
\Sigma_B(s';\theta) &= \gamma \, \mathrm{diag}\!\bigl(\exp \{C_B\tanh(w_B(s';\theta))\}\bigr),
\end{aligned}
\end{equation}
where \(u_B,w_B:\mathbb R^d\to\mathbb R^d\) are learnable functions and $C_B=4$. Here, we employ a homogeneous kernel, as our formulation does not rely on an explicit notion of time. The backward density is inspired by an Ornstein--Uhlenbeck-type process targeting a zero-mean distribution, effectively acting as a mechanism for destructing or reversing the forward generative process, similar in spirit to diffusion samplers \citep{zhang2021path}. We further introduce a learned correction $u_B$ to the drift of the backward kernel, following \citep{gritsaev2025adaptive}, where the backward process is enhanced by learnable adjustments to both its drift and variance.

We now justify why \cref{ass:finite_exp_len} holds in our setting in practice. Provided that $p = \inf_{s \in \mathbb{R}^d} d_F(s) > 0$, we have $\mathbb{E}_{\tau \sim P}[n_{\tau}] \le \mathbb{E}[\xi] = 1/p < \infty$, where $\xi \sim \operatorname{Geom}(p)$. We impose a lower-bound for a classifier $d_F(s) \ge \varepsilon$ for any $s \in \mathbb{R}^d$, where $\varepsilon = 10^{-5}$.

\subsection{Target densities}
\label{sec:envs}

We evaluate the proposed method on a set of synthetic benchmark distributions commonly used to assess sampling algorithms in continuous spaces. These environments are designed to capture different challenges such as multimodality or complex geometry.

\paragraph{GMM9.}
The GMM9 environment is a Gaussian mixture model with nine components arranged randomly in [-5,5]. Each component has diagonal covariance and the mixture weights are non-uniform, making the distribution both multimodal and imbalanced. Formally, the density is given by a mixture of Gaussian components
\[
\pi(x) = \sum_{k=1}^{9} w_k \, \mathcal{N}(x \mid \mu_k, \Sigma_k),
\]
where $\Sigma_k = 0.55 \, I_d$ and $\sum_{k=1}^9 w_k = 1$. This environment tests the ability of the sampler to correctly capture multiple modes and their relative probabilities.

\paragraph{Funnel.}
The Funnel distribution is a hierarchical model with strong scale variation across dimensions. The first coordinate \(x_1\) controls the variance of the remaining coordinates:
\[
x_1 \sim \mathcal{N}(0, 3^2), \qquad
x_{2:d} \sim \mathcal{N}(0, \exp(x_1) I).
\]
This leads to regions of very different scales, creating a narrow ``funnel'' structure. Such distributions are well known to be challenging for MCMC methods due to poor mixing between regions of different variance.

\paragraph{ManyWell.}
The ManyWell density consists of multiple independent double-well potentials, leading to an exponentially large number of modes in higher dimensions. The energy is constructed as a sum of double-well energies:
\[
U(x_1, x_2) = a x_1 + b x_1^2 + cx_1^4 + \frac{x_2^2}{2},
\]
which induces a multimodal distribution with well-separated modes, where we set $a=-0.5,b=-6.0,c=1.0$. This environment is particularly challenging because it requires efficient transitions between a large number of modes.
\paragraph{Digits.}
The Digits environment is adopted from \cite{blessing2024beyond} and corresponds to sampling MNIST images from a density defined by a NICE normalizing flow model \cite{dinh2014nice}. The target distribution is a continuous density over \(\mathbb R^{196}\), obtained by training an invertible flow on \(14\times14\) MNIST images. Although the underlying dataset consists of images, the resulting target distribution has full support on \(\mathbb R^{196}\). Samples from this environment therefore correspond to continuous vectors that can be reshaped into grayscale digit images. This task provides a substantially more challenging high-dimensional benchmark than the synthetic multimodal targets considered in the main text.
\subsection{Algorithm details and hyperparameters}
\label{sec:main_alg_details}
We train the proposed non-acyclic GFlowNet using the prefix trajectory balance objective described in \cref{sec:main_algorithm}. The overall training procedure is summarized in \cref{alg:non_acyclic_training}. In this section, we provide additional implementation details.

At each iteration, the algorithm alternates between forward (on-policy) and backward (off-policy) updates. During forward updates, trajectories are sampled from the forward policy starting from the initial distribution \(p_0\). During backward updates, terminal states are sampled from a replay buffer and trajectories are generated using the backward policy. This alternating scheme allows the model to combine on-policy exploration with off-policy refinement and was inspired by \citep{sendera2024improved}. Across all experiments, we used a replay ratio of $N_{\text{RR}}=2$, which specifies the number of off-policy iterations performed within each on-policy iteration.

Practically, collecting data with early termination hinders exploration of the continuous space. This issue motivates us to disable classifiers during trajectory collection and instead sample trajectories of a fixed length $N_{\max}$. We choose $N_{\max}$ based on the number of iterations required for ULA with the same step-size $\gamma$ to reach a stationary regime, monitored via the Wasserstein metric. If $N_{\max}$ is chosen too small, trajectories may fail to reach proper high-density regions, resulting in poor exploration of the state space. Conversely, setting a very large $N_{\max}$ slows down training and is impractical, though it remains functional because we apply flow regularization to encourage shorter trajectories.

Following \citep{sendera2024improved}, to improve sample efficiency, we employ a prioritized replay buffer that stores terminal states. Each state is assigned a priority based on its reward, which biases sampling toward high-reward regions of the state space. The buffer is updated continuously as new samples are generated from the forward policy. Before training, we pre-fill the buffer with $100 \times B$ terminal samples from a forward policy initialized as Langevin dynamics, where $B$ is the batch size. This initial pre-filling improves training at the beginning.

We further incorporate a local search procedure based on the Metropolis-adjusted Langevin algorithm (MALA) \citep{sendera2024improved}. At regular intervals $N_{\text{LS}}$, states sampled from the replay buffer are refined using a small number of MALA steps, and the resulting samples are added back to the buffer. This improves exploration by enabling transitions between distant high-density regions and helps mitigate the limitations of purely learned dynamics. In our implementation, we used 300 MALA steps with $\gamma=10^{-3}$ and set $N_{\text{LS}}=25$.

The forward and backward policies are parameterized by neural networks as described in \cref{sec:param}. We use a shared backbone consisting of a 3-layer MLP for forward and backward embeddings, followed by a 2-layer MLP for their respective specific predictions, with a hidden dimension of 128. The forward continuous kernel $q_F$ is either fixed (in the ULA-based setup) or augmented with learned drift and variance corrections, while the backward kernel is always learned. The forward and backward classifiers control the stopping behavior and are trained jointly with the transition kernels. The normalizing constant \(\log Z_\theta\) is also learned as a scalar parameter with a learning rate \(10^{-1}\). We set the variance of the initial distribution $\sigma^2_0=1$ for low-dimensional environments and $\sigma^2_0=5$ for high-dimensional ones.

Optimization is performed using Adam with a batch size of 2000 with gradient clipping. Environment-specific hyperparameters are summarized in the next sections, where we discuss experiments in details. All experiments were performed on NVIDIA A100 GPUs. Our implementations are based upon the published code of~\cite{blessing2024beyond}.

\begin{algorithm}[tb]
  \caption{Training non-acyclic continuous GFlowNet with prefix trajectory balance}
  \label{alg:non_acyclic_training}
  \begin{algorithmic}
    \STATE {\bfseries Input:} Initial parameters $\theta$, reward $R$, initial distribution $p_0$, maximum trajectory length $N_{\max}$, batch size $B$, number of training steps $K$, replay buffer $\mathcal B$, regularization coefficient $\rho$, replay ratio $N_{\text{RR}}$, local search frequency $N_{\text{LS}}$
    \STATE {\bfseries Output:} Trained model parameters $\theta$
    \STATE Pre-fill buffer $\mathcal B$ with terminal states from $p_F$
    \FOR{$k=1$ {\bfseries to} $K$}
        \IF{$(k-1) \mod N_{\text{RR}} \equiv 0$}
            \STATE Sample initial states $\{s_1^b\}_{b=1}^B \sim p_0$
            \STATE Sample trajectories $\widetilde \tau=\{(s_0\to s_1^b\to \cdots \to s_{N_{\max}}^b=x^b)\}_{b=1}^B$ using $p_F$ with $d_F$ disabled
            \STATE Add $\{x^b\}_{b=1}^B$ to $\mathcal B$ with priorities $r(x^b)$
        \ELSE
            \STATE Sample terminal states $\{x^b\}_{b=1}^B \sim \mathcal B$
            \STATE Sample trajectories $\widetilde \tau=\{(s_0\to s_1^b\to \cdots \to s_{N_{\max}}^b=x^b)\}_{b=1}^B$ using $p_B$ with $d_B$ disabled
        \ENDIF
        \STATE Set $\tau^b_{0:i} = \widetilde \tau^b_{0:i} \cup \{ s_f \}$ for $i=1,\dots,N_{\max},b=1,\dots,B$
        \STATE For all complete prefixes $\{\tau^b_{0:i}\}_{b=1,i=1}^{B,N_{\max}}$, compute TB loss $\mathcal L_{\mathrm{TB}} (\theta,\tau^b_{0:i})$ and weights $\{w^b_i\}_{b=1}^B$
        \STATE Compute
        \[
        \mathcal L(\theta)
        =
        \frac{1}{B}
        \sum_{b=1}^B
        \sum_{i=1}^{N_{\max}}
        w_i^b
        \left[
        \mathcal L_{\mathrm{TB}}(\theta,\tau^b_{0:i})
        +
        \rho\frac{r(s_i^b)}{d_F(s_i^b;\theta)}
        \right]
        \]
        \STATE Update $\theta$ by a gradient step on $\mathcal L(\theta)$
        \IF{$(k-1) \mod N_{\text{LS}} \equiv 0$}
            \STATE Sample $\{z^b\}_{b=1}^B\sim\mathcal B$, run MALA to obtain $\{\bar z^b\}_{b=1}^B$, and add $\{\bar z^b\}_{b=1}^B$ to $\mathcal B$
        \ENDIF
    \ENDFOR
  \end{algorithmic}
\end{algorithm}
\subsection{Details on the improvement over ULA}
\label{sec:appendix_exp_improve_ula}
As additional stopping-based baselines, we consider ULA with the Gelman--Rubin convergence diagnostic (ULA RG) and ULA with the Geweke convergence diagnostic (ULA Geweke). Formal definitions of the corresponding diagnostics are provided in \cref{appendix:mcmc-stopping}. For ULA RG, we run several parallel ULA chains and evaluate the multivariate $\hat R$ statistic on a predefined grid of trajectory lengths $t$ using only trajectory prefixes of length $t$. Sampling is stopped at the first checkpoint where the maximum $\hat R$ value across coordinates falls below a fixed threshold. In our experiments, we use a threshold of $1.1$. If the criterion is never satisfied before the final checkpoint, the trajectory is truncated at the maximal allowed length. For ULA Geweke, we independently analyze each trajectory by comparing statistics of an early trajectory window (first $10\%$ of samples) and a late window (last $50\%$ of samples) within the prefix of length $t$. A Geweke $Z$-score is computed for each coordinate, and the stopping time is defined as the first checkpoint where the maximum absolute $Z$-score across coordinates falls below a threshold of $4.0$. Unlike ULA RG, the stopping time for ULA Geweke is determined independently for each chain.

In \cref{tab:params_improved_ula} we specify hyperparameters for the runs in \cref{tab:improve_ula_table_main}.
\begin{table}[h!]

\centering
\caption{\label{tab:params_improved_ula}Environment-specific hyperparameters for ULA experiments, where lr is a learning rate of the optimizer, $\rho$ is a regularization coefficient, $\gamma$ is the step-size, $N_{\max}$ is trajectory length during training and grad clip specifies values of the gradient clipping with respect to parameters in the optimizer.}
\begin{tabular}{@{}lccccc}
\toprule
Environment $\downarrow$ Parameter $\rightarrow$ & lr & $\rho$ & $\gamma$ & $N_{\max}$ & grad clip \\
\midrule
GMM9 ($d=2$)     & $10^{-3}$ & $10^{-1}$ & $10^{-1}$ & $10^2$ & $10^{0}$ \\
Funnel ($d=2$)   & $10^{-4}$ & $10^{-2}$ & $10^{-2}$ & $10^2$ & $10^{-3}$ \\
GMM9 ($d=40$)    & $10^{-4}$ & $10^{0}$ & $10^{-1}$ & $10^2$ & $10^{-3}$ \\
ManyWell ($d=40$) & $10^{-5}$ & $10^{-1}$ & $10^{-2}$ & $10^2$ & $10^{-5}$ \\
\bottomrule
\end{tabular}
\end{table}

\subsection{Details on the multilevel scheme experiments}
\label{sec:appendix_exp_ml}
We added the positional encoding \citep{vaswani2017attention} for the levels variables in neural network. Then a state $s$ is concatenated with this level embedding and passed to the MLP. For a schedule $\beta_l$ we adapt a linear schedule from $0$ to $1$.
In \cref{tab:params_ml} we specify hyperparameters for the runs in \cref{tab:ml_table_main}.
\begin{table}[h!]
\centering
\caption{\label{tab:params_ml}Environment-specific hyperparameters for multi-level experiments, where lr is a learning rate of the optimizer, $\rho$ is a regularization coefficient, $\gamma$ is the step-size, $N_{\max}$ is trajectory length during training and grad clip specifies values of the gradient clipping with respect to parameters in the optimizer.}
\begin{tabular}{@{}lccccc}
\toprule
Environment $\downarrow$ Parameter $\rightarrow$ & lr & $\rho$ & $\gamma$ & $N_{\max}$ & grad clip \\
\midrule
GMM9 ($d=40$)    & $10^{-4}$ & $10^{0}$ & $10^{-1}$ & $10^2$ & $10^{-3}$ \\
ManyWell ($d=40$) & $10^{-5}$ & $10^{-1}$ & $10^{-2}$ & $10^2$ & $10^{-5}$ \\
Digits ($d=196$)    & $10^{-4}$ & $10^{0}$ & $10^{-1}$ & $5 \times 10^1$ & $10^{-3}$ \\
\bottomrule
\end{tabular}
\end{table}

\subsection{Metrics}
\label{sec:metrics}

Let \(\pi(x)=r(x)/Z\) denote the target distribution, and let \(p_T\) be the marginal distribution of terminal states produced by the model. Following \citep{blessing2024beyond}, to assess the discrepancy between \(p_T\) and \(\pi\), we report  the following sample-based metrics that are widely used in the evaluation of generative samplers. Additionally, to measure performance of our method we report ELBO and EUBO \citep{blessing2024beyond}. For these metrics we utilize the implementation from \url{https://github.com/DenisBless/variational_sampling_methods}. Furthermore, we calculate the mean trajectory length of our sampler and denote this quantity by $\overline{n_{\tau}}$.

\subsubsection*{Maximum Mean Discrepancy}
Maximum Mean Discrepancy \citep{gretton2012kernel, blessing2024beyond} is a kernel-based integral probability metric that compares two distributions through their embeddings in a reproducing kernel Hilbert space (RKHS). Let \(\mathcal H_k\) be the RKHS associated with a positive definite kernel \(k\). The MMD between \(p_T\) and \(\pi\) is defined as
\begin{equation}
\operatorname{MMD}_k(p_T,\pi)
=
\sup_{\substack{f\in\mathcal H_k\\ \|f\|_{\mathcal H_k}\le 1}}
\left(
\mathbb E_{x\sim p_T}[f(x)]-\mathbb E_{y\sim \pi}[f(y)]
\right).
\end{equation}
For characteristic kernels, this quantity is nonnegative and vanishes if and only if \(p_T=\pi\). Given samples \(X=\{x_i\}_{i=1}^n \sim p_T\) and \(Y=\{y_j\}_{j=1}^m \sim \pi\), we use the usual unbiased estimator of the squared MMD:
\begin{equation}
\widehat{\operatorname{MMD}}_k^2(X,Y)
=
\frac{1}{n(n-1)}\sum_{i\neq j} k(x_i,x_j)
+
\frac{1}{m(m-1)}\sum_{i\neq j} k(y_i,y_j)
-
\frac{2}{nm}\sum_{i=1}^n\sum_{j=1}^m k(x_i,y_j).
\end{equation}
In our experiments, the squared exponential kernel
\begin{equation}
k(x,y)=\exp\!\left(-\frac{\|x-y\|_2^2}{\alpha}\right)
\end{equation}
is used with bandwidth \(\alpha\) selected by the median heuristic \citep{blessing2024beyond}.

\subsubsection*{Entropic optimal transport distance}

As a complementary metric, we use an entropy-regularized optimal transport distance \citep{peyre2019computational, blessing2024beyond} based on the quadratic cost. The underlying unregularized quantity is the 2-Wasserstein distance,
\begin{equation}
W_{2,\varepsilon}(p_T,\pi)
=
\inf_{\xi \in \Gamma(p_T,\pi)}
\left(
\int_{\mathbb R^d\times\mathbb R^d}
\|x-y\|_2^2\,\xi(dx,dy) - \epsilon \mathcal H(\xi)
\right)^{1/2},
\end{equation}
where \(\Gamma(p_T,\pi)\) denotes the set of couplings with marginals \(p_T\) and \(\pi\), and $\mathcal H(\xi) = - \int_{\mathbb{R}^d \times \mathbb{R}^d} \xi(x,y) \log \xi(x,y) \, \mathrm dx \mathrm dy$. In practice, we compute its entropically regularized counterpart using Sinkhorn iterations, which is substantially more tractable numerically. Throughout all experiments, we set \(\epsilon=10^{-3}\) \citep{blessing2024beyond}.
\subsubsection*{Evidence lower and upper bounds}
For a given terminal point $x \in \mathbb{R}^d$, let $\mathcal T_{x} = \{ \tau \in \mathcal T \mid s_{n_{\tau}} = x\}$. For $\tau \in \mathcal T_{x}$ one can write trajectory distributions induced by the transition densities $p_F$ and $p_B$ from \cref{sec:main_algorithm} as $p_F(\tau)$ and $p_B(\tau) = p_B(\tau \mid s_{n_\tau} = x) \pi(x)$.  then we have:
\begin{equation}
\begin{aligned}
\label{eq:z}
Z &= \int_{\mathbb{R}^d} r(x) dx = \int_{\mathbb{R}^d} r(x) \bigg(\int_{\mathcal T_{x}} p_B(\tau \mid s_{n_\tau} = x)  d \tau \bigg) dx\\
&= \int_{\mathcal T} r(x) p_B(\tau \mid s_{n_\tau} = x) d \tau = \mathbb{E}_{\tau \sim p_F}\bigg[ \frac{r(x) p_B(\tau \mid s_{n_\tau} = x)}{p_F(\tau)} \bigg].
\end{aligned}
\end{equation}
One way to construct an estimator for $\log Z$ is to use the evidence lower-bound (ELBO) \citep{blessing2024beyond}:
\begin{equation}
\log Z = \log \mathbb{E}_{\tau \sim p_F}\bigg[ \frac{r(x) p_B(\tau \mid s_{n_\tau} = x)}{p_F(\tau)} \bigg] \ge \mathbb{E}_{\tau \sim p_F}\bigg[\log r(x) + \log p_B(\tau \mid s_{n_\tau} = x) - \log p_F(\tau)  \bigg].
\end{equation}
To estimate the ELBO, we draw $M=2000$ samples from the current policy and compute the average of the estimated $\log Z$ values:
\begin{equation}
\text{ELBO} = \frac{1}{M} \sum_{i=1}^M \bigg[\log r(x^i) + \log p_B(\tau^i \mid s_{n_\tau} = x^i) - \log p_F(\tau^i) \bigg], \quad \tau^i \sim p_F, \tau^i \leadsto x^i.
\end{equation}
To measure mode coverage, evidence upper-bound (EUBO) metric \citep{blessing2024beyond} was introduced, but it requires samples from $\pi$:
\begin{equation}
\begin{aligned}
\log Z &= \mathbb{E}_{x \sim \pi, \tau \sim p_B(\cdot \mid s_{n_\tau} = x)} \bigg[\log \frac{ r(x) p_B(\tau \mid s_{n_\tau} = x)}{p_F(\tau)} \bigg] \\
&- \mathbb{KL}(\pi \cdot p_B \mid \mid p_F) \le \mathbb{E}_{x \sim \pi, \tau \sim p_B(\cdot \mid x)} \bigg[\log \frac{ r(x) p_B(\tau \mid s_{n_\tau} = x)}{p_F(\tau)} \bigg].
\end{aligned}
\end{equation}
We draw $M=2000$ samples from $\pi$ and then simulate trajectories using a backward policy:
\begin{equation}
\text{EUBO} = \frac{1}{M} \sum_{i=1}^M \bigg[\log r(x^i) + \log p_B(\tau^i \mid s_{n_\tau} = x^i) - \log p_F(\tau^i) \bigg], \quad x^i \sim \pi, \tau^i \sim p_B(\cdot \mid s_{n_\tau} = x^i).
\end{equation}

%\newpage
\subsection{Additional visualizations}
\label{sec:add_vis}

We provide qualitative sample visualizations for all environments studied in the experiments in \cref{sec:exp_improve_ula}. For each environment, we show samples generated by ULA, ULA RG, ULA Geweke, ULA with a learned classifier, and the full model.

\begin{figure*}[h!]
\centering
% first row
\begin{subfigure}{0.30\textwidth}
    \centering
    \includegraphics[width=\linewidth]{figures/samples/gmm9_2d_ula.png}

    \small ULA
\end{subfigure}
\hfill
\begin{subfigure}{0.30\textwidth}
    \centering
    \includegraphics[width=\linewidth]{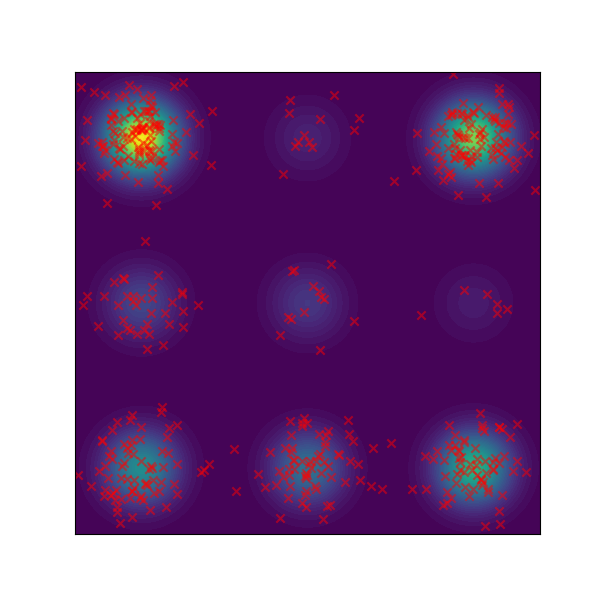}

    \small ULA RG
\end{subfigure}
\hfill
\begin{subfigure}{0.30\textwidth}
    \centering
    \includegraphics[width=\linewidth]{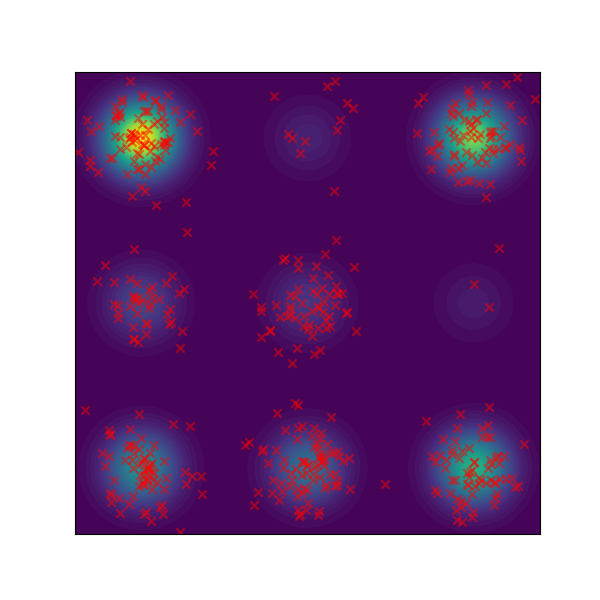}

    \small ULA Geweke
\end{subfigure}
\vspace{0.5em}
% second row
\begin{center}
\begin{subfigure}{0.30\textwidth}
    \centering
    \includegraphics[width=\linewidth]{figures/samples/gmm9_2d_ula_clf.png}

    \small ULA + clf
\end{subfigure}
\hspace{0.05\textwidth}
\begin{subfigure}{0.30\textwidth}
    \centering
    \includegraphics[width=\linewidth]{figures/samples/gmm9_2d_full_model.png}

    \small ULA w/ clf \& corr
\end{subfigure}
\end{center}

\caption{Samples on 9GMM ($d=2$) with step-size $\gamma=0.1$.}
\label{fig:samples_gmm9_2d_app}
\end{figure*}

\begin{figure*}[h!]
\centering

% first row
\begin{subfigure}{0.30\textwidth}
    \centering
    \includegraphics[width=\linewidth]{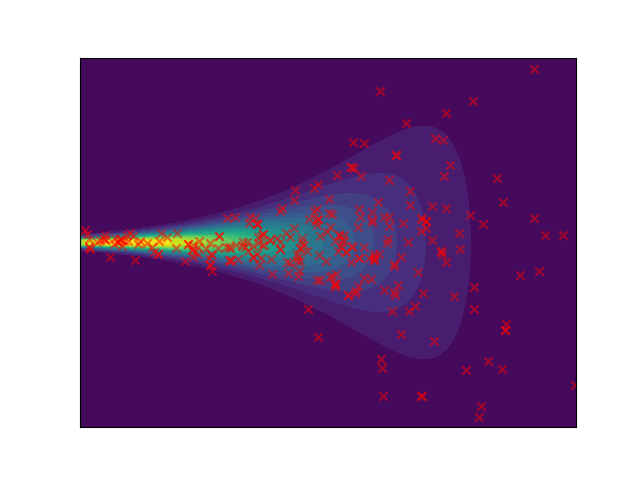}

    \small ULA
\end{subfigure}
\hfill
\begin{subfigure}{0.30\textwidth}
    \centering
    \includegraphics[width=\linewidth]{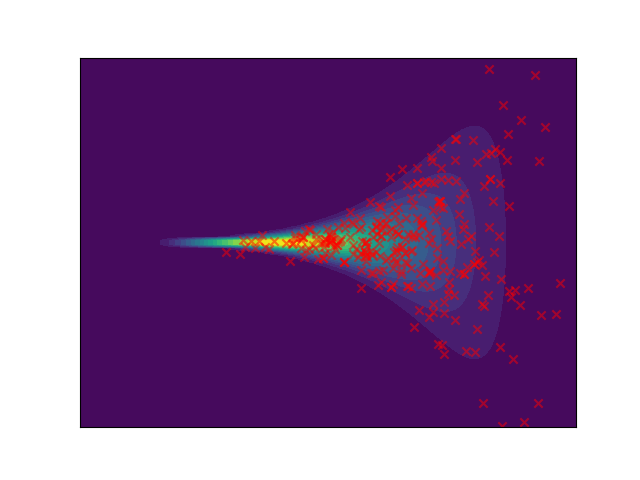}

    \small ULA RG
\end{subfigure}
\hfill
\begin{subfigure}{0.30\textwidth}
    \centering
    \includegraphics[width=\linewidth]{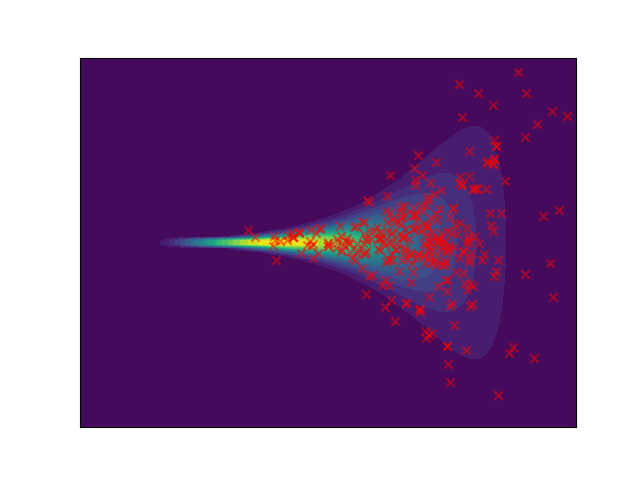}

    \small ULA Geweke
\end{subfigure}

\vspace{0.5em}

% second row
\begin{center}
\begin{subfigure}{0.30\textwidth}
    \centering
    \includegraphics[width=\linewidth]{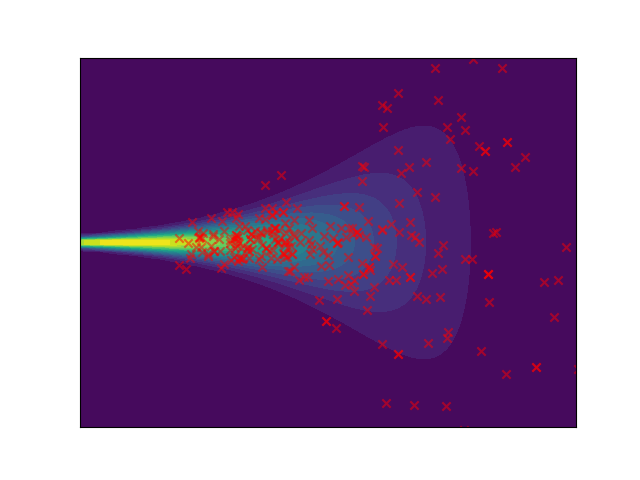}

    \small ULA + clf
\end{subfigure}
\hspace{0.05\textwidth}
\begin{subfigure}{0.30\textwidth}
    \centering
    \includegraphics[width=\linewidth]{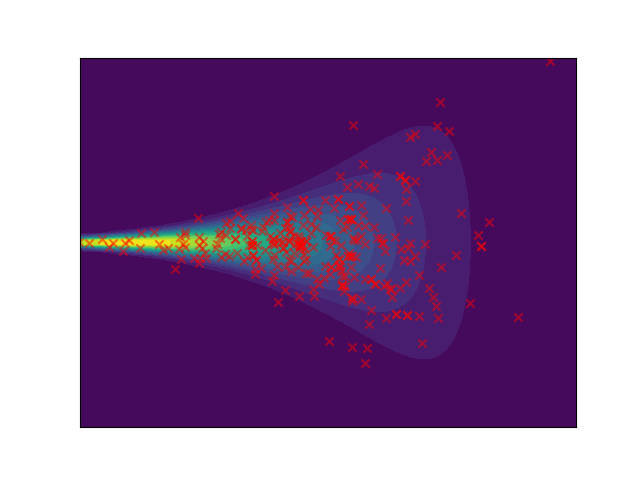}

    \small ULA w/ clf \& corr
\end{subfigure}
\end{center}

\caption{Samples on Funnel ($d=2$) with step-size $\gamma=0.01$.}
\label{fig:samples_funnel_2d}
\end{figure*}

\begin{figure*}[h!]
\centering

% first row
\begin{subfigure}{0.30\textwidth}
    \centering
    \includegraphics[width=\linewidth]{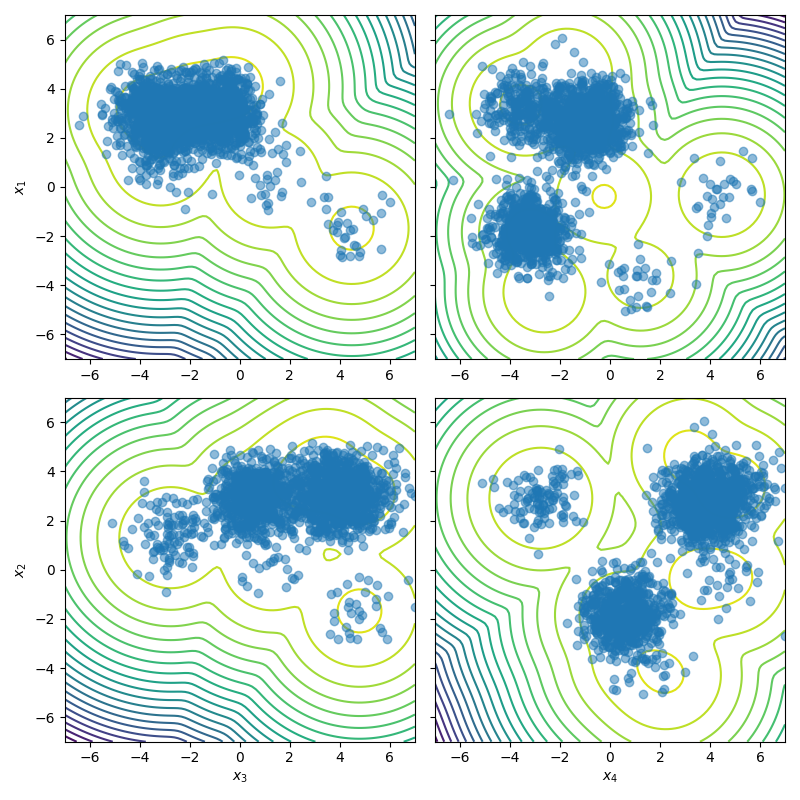}

    \small ULA
\end{subfigure}
\hfill
\begin{subfigure}{0.30\textwidth}
    \centering
    \includegraphics[width=\linewidth]{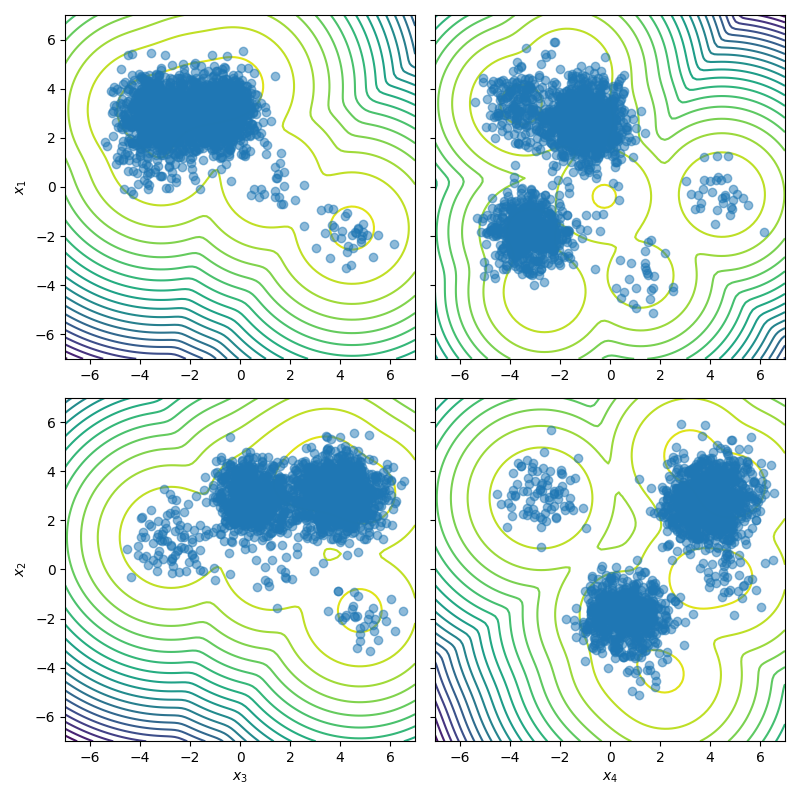}

    \small ULA RG
\end{subfigure}
\hfill
\begin{subfigure}{0.30\textwidth}
    \centering
    \includegraphics[width=\linewidth]{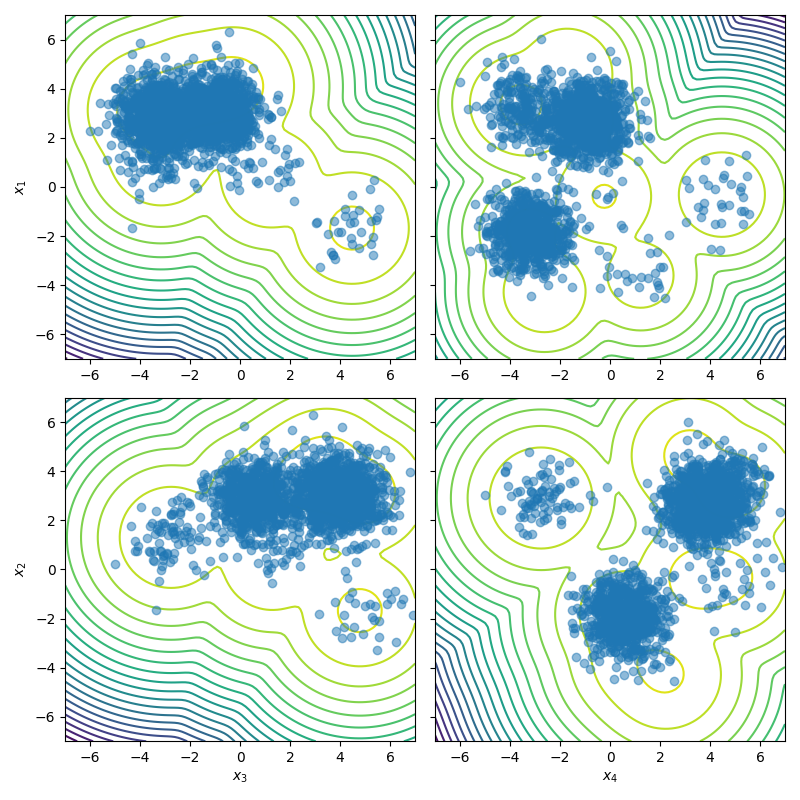}

    \small ULA Geweke
\end{subfigure}

\vspace{0.5em}

% second row
\begin{center}
\begin{subfigure}{0.30\textwidth}
    \centering
    \includegraphics[width=\linewidth]{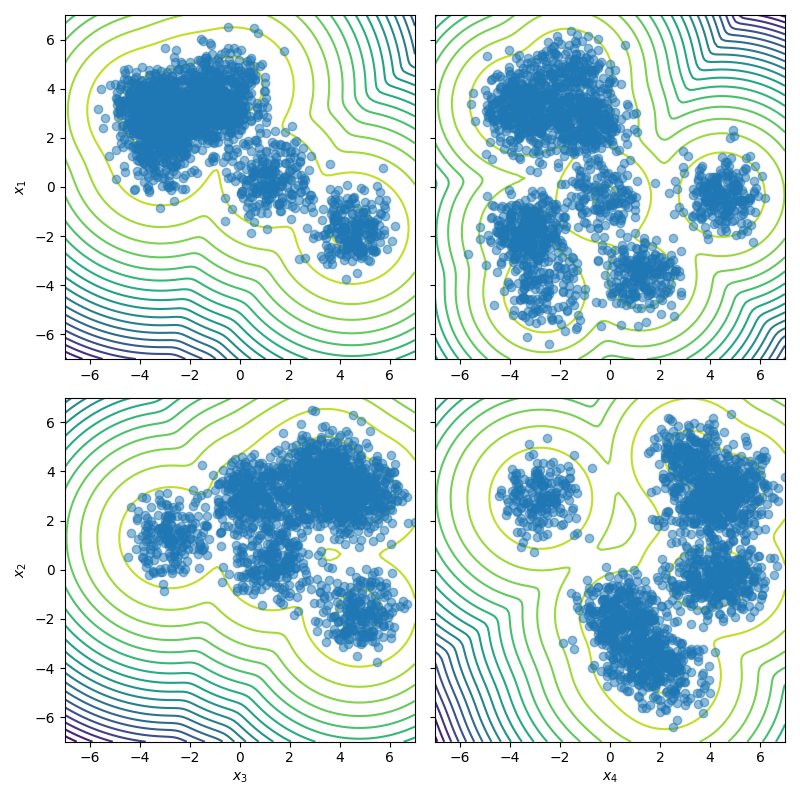}

    \small ULA + clf
\end{subfigure}
\hspace{0.05\textwidth}
\begin{subfigure}{0.30\textwidth}
    \centering
    \includegraphics[width=\linewidth]{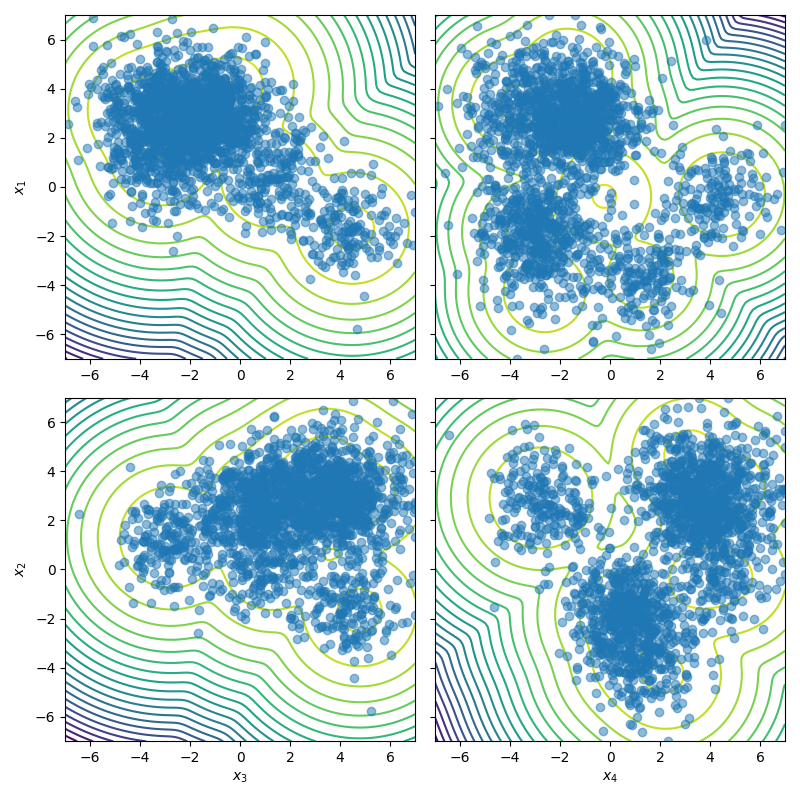}

    \small ULA w/ clf \& corr
\end{subfigure}
\end{center}

\caption{Samples on GMM9 ($d=40$) using marginal projections, step-size $\gamma=0.1$.}
\label{fig:samples_gmm9_40d}
\end{figure*}

\begin{figure*}[h!]
\centering

% first row
\begin{subfigure}{0.30\textwidth}
    \centering
    \includegraphics[width=\linewidth]{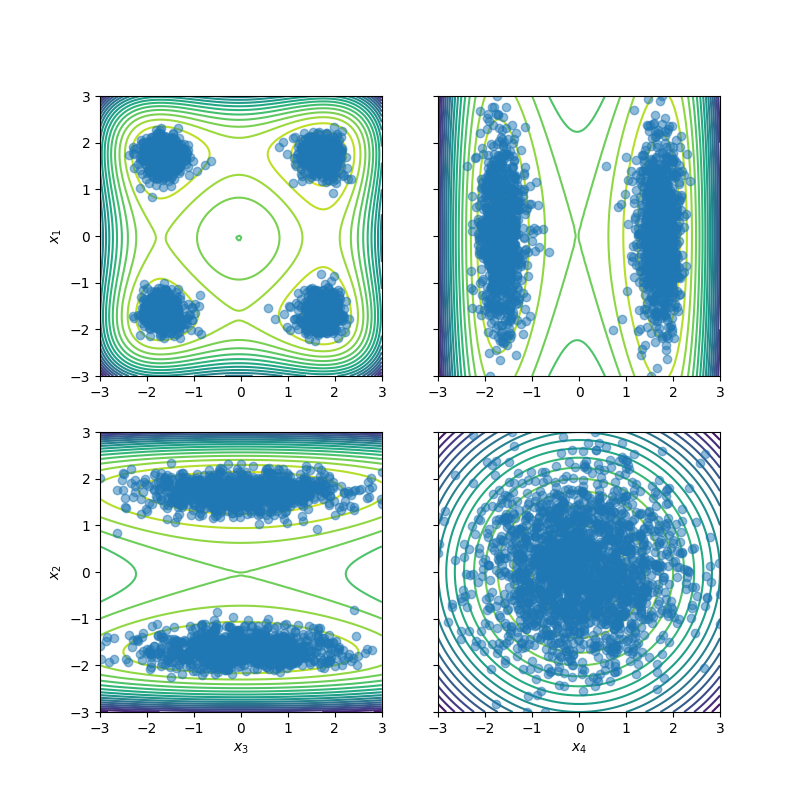}

    \small ULA
\end{subfigure}
\hfill
\begin{subfigure}{0.30\textwidth}
    \centering
    \includegraphics[width=\linewidth]{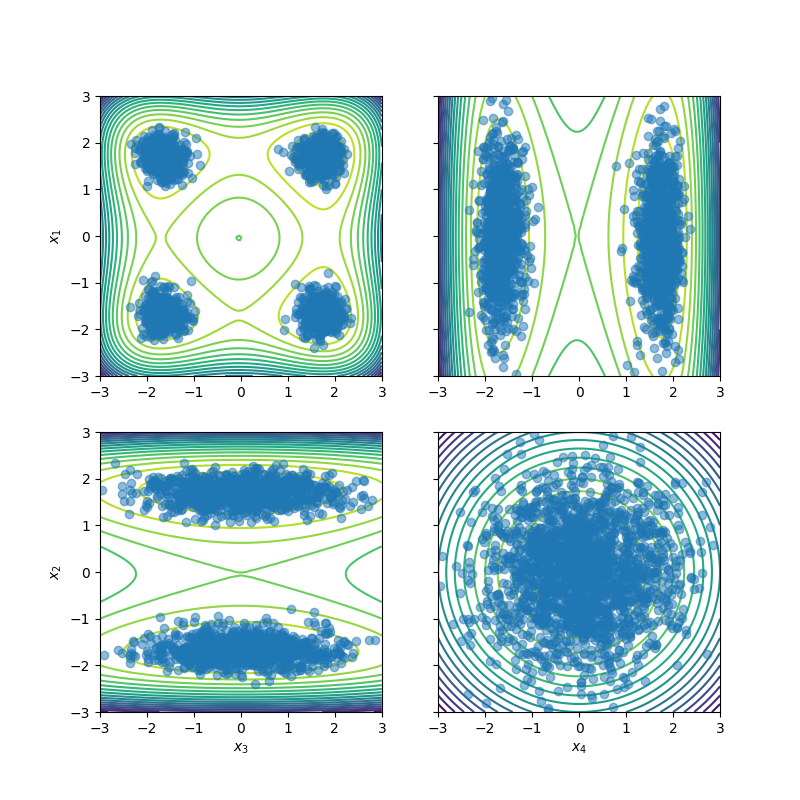}

    \small ULA RG
\end{subfigure}
\hfill
\begin{subfigure}{0.30\textwidth}
    \centering
    \includegraphics[width=\linewidth]{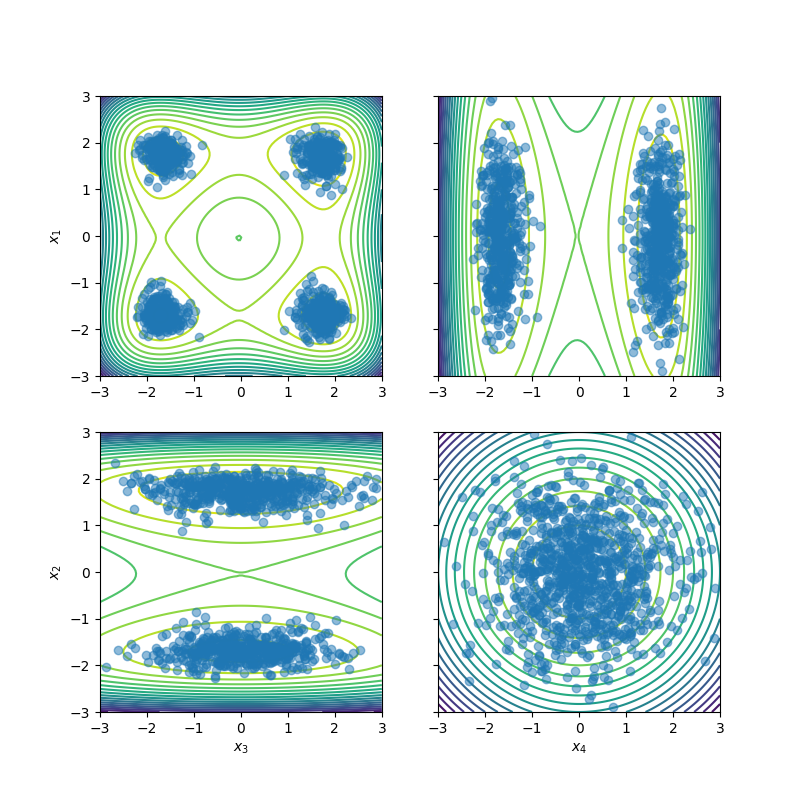}

    \small ULA Geweke
\end{subfigure}

\vspace{0.5em}

% second row
\begin{center}
\begin{subfigure}{0.30\textwidth}
    \centering
    \includegraphics[width=\linewidth]{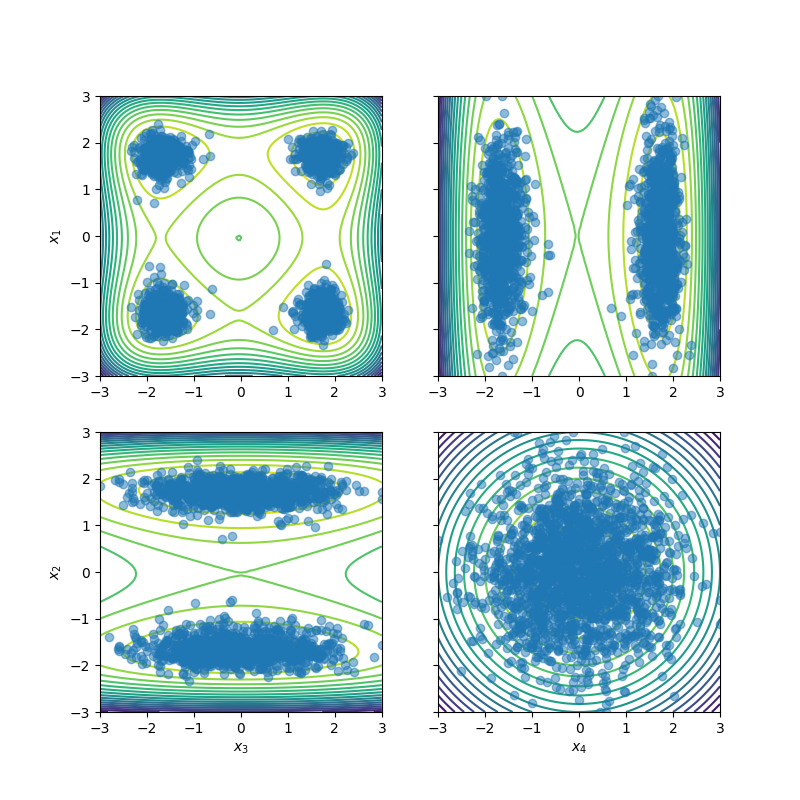}

    \small ULA + clf
\end{subfigure}
\hspace{0.05\textwidth}
\begin{subfigure}{0.30\textwidth}
    \centering
    \includegraphics[width=\linewidth]{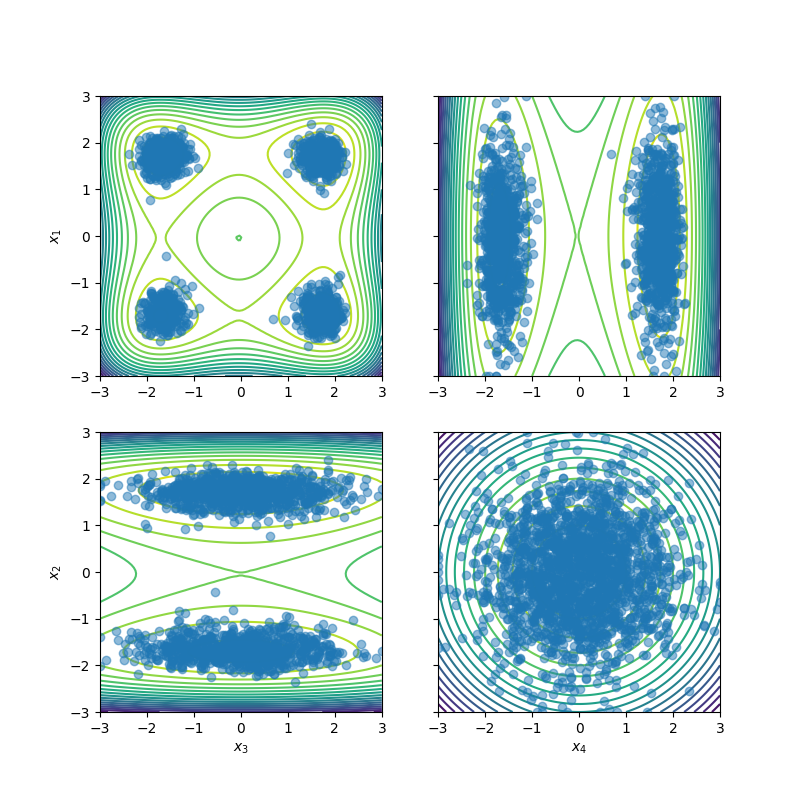}

    \small ULA w/ clf \& corr
\end{subfigure}
\end{center}

\caption{Marginal projections of samples from ManyWell ($d=40$), step-size $\gamma=0.1$.}
\label{fig:samples_many_well_40d}
\end{figure*}

\clearpage

\begin{figure*}[h!]
\centering
\includegraphics[width=0.32\textwidth]{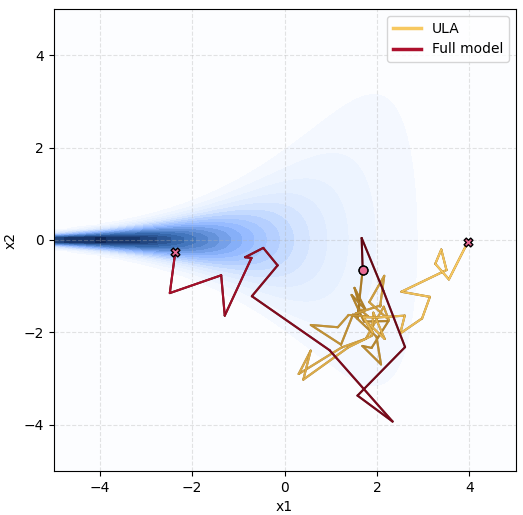}
\includegraphics[width=0.32\textwidth]{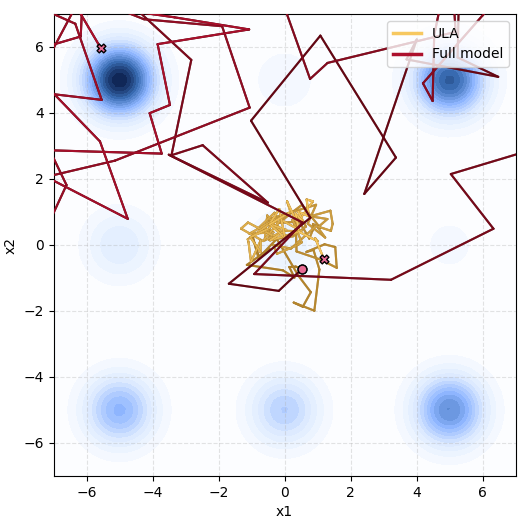}
\includegraphics[width=0.32\textwidth]{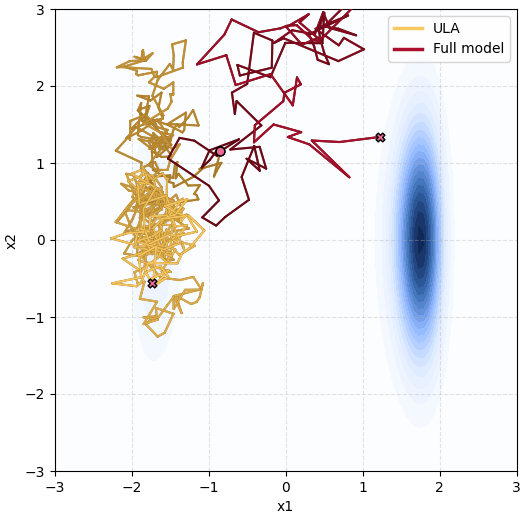}
\caption{
Comparison of trajectories produced by ULA and the full model in different environments (from left to right: Funnel, GMM9, ManyWell).
ULA tends to remain trapped within a single mode or local region of the target and therefore requires a large mixing time to move between modes.
In contrast, the proposed model is able to make effective transitions between modes and more distant high-density regions, resulting in substantially more efficient exploration of the target distribution.
}
\label{fig:improve_ula_traj_compare}
\end{figure*}

\section{Classical stopping diagnostics for MCMC}
\label{appendix:mcmc-stopping}

For $m$ chains of length $n$, let $\bar x_j$ be the mean of chain $j$, let
\[
\bar x = \frac{1}{m}\sum_{j=1}^m \bar x_j
\]
denote the overall mean across chains, and let $s_j^2$ be the within-chain variance. The Gelman--Rubin potential scale reduction factor is based on
\begin{equation}
B = \frac{n}{m-1}\sum_{j=1}^m (\bar x_j - \bar x)^2,
\qquad
W = \frac{1}{m}\sum_{j=1}^m s_j^2,
\end{equation}
and
\begin{equation}
\hat R = \sqrt{\frac{\hat V}{W}},
\qquad
\hat V = \frac{n-1}{n}W + \frac{1}{n}B.
\label{eq:rhat}
\end{equation}
Values of $\hat R$ close to one indicate approximate agreement between between-chain and within-chain variability.

Another commonly used quantity is the effective sample size,
\begin{equation}
\mathrm{ESS} = \frac{N}{1 + 2\sum_{k\ge 1}\rho_k},
\label{eq:ess}
\end{equation}
where $N$ is the total number of retained samples and $\rho_k$ denotes the lag-$k$ autocorrelation.

A related diagnostic is Geweke's statistic,
\begin{equation}
Z = \frac{\bar g_A - \bar g_B}{\sqrt{\hat S_A(0)/n_A + \hat S_B(0)/n_B}},
\label{eq:geweke}
\end{equation}
which compares the averages of a scalar summary over an early window $A$ and a late window $B$, with $\hat S_A(0)$ and $\hat S_B(0)$ denoting corresponding long-run variance estimates.

\section{Interpretation of \cref{thm:conn_mcmc}}
\label{sec:conn_mcmc_interpret}
In this section we discuss and interpret results of \cref{thm:conn_mcmc}. At first, let us define useful objects for this matter. Following \cite{brunswic2024theory}, given a flow measure $F$ from \cref{def:flow}, we define an edge flow
\begin{equation}
    F(A \times B) = \int_A F(\mathrm ds) P_F(s, B),
\end{equation}
for any measurable $A \subseteq \bar{\mathcal{S}} \setminus \{ s_f\}, B \subseteq \bar{\mathcal{S}} \setminus \{ s_0\}$. Then using an edge flow we can define an incoming and outcoming flow measures $F_{\text{in}}$ and $F_{\text{out}}$, respectively:
\begin{equation}
\begin{aligned}
F_{\text{in}}(B) &= F(\bar{\mathcal{S}} \setminus \{ s_f\} \times B),\\
F_{\text{out}}(A) &= F(A \times \bar{\mathcal{S}} \setminus \{ s_0\}).
\end{aligned}
\end{equation}
Consider a measurable set $A \subseteq \mathbb{R}^d$. Then from the definitions of transitions in our state space
\begin{equation}
\begin{aligned}
    P_F(s_0, A) &= p_0(A),\\
    P_F(s, A) &= (1 - d_F(s)) Q_F(s, A) + d_F(s) \delta_{s_f}(A),
\end{aligned}
\end{equation}
we obtain
\begin{equation}
\begin{aligned}
F_{\text{in}}(A) &= \int_{\mathcal{S} \cup \{s_0\}} F(\mathrm ds) P_F(s, B) = F(\{s_0\}) p_0(A) + \int_{\mathbb{R}^d} F(\mathrm ds) (1 - d_F(s)) Q_F(s, A)\\
&= Z p_0(A) + \int_{\mathbb{R}^d} (F(\mathrm ds) - R(\mathrm ds)) Q_F(s, A) = Z p_0(A) - (F - R)Q_F(A),
\end{aligned}
\end{equation}
where we used that from \cref{prop:flow_yiels_fm} the flow measure $F$ and the Markov kernel $P_F$ satisfy flow matching conditions \eqref{eq:fm}. Therefore, detailed balance conditions \eqref{eq:db} must hold and we can express a classifier $d_F$ from a relation \eqref{eq:flow_reward_clf}. In turn, a flow measure we defined in \cref{def:flow} coincides with the outcoming flow measure $F_{\text{out}}$:
\begin{equation}
\begin{aligned}
F_{\text{out}}(A) = \int_{A} F(\mathrm ds) P_F(s, \bar{S} \setminus \{ s_0 \}) = \int_{A} F(\mathrm ds) = F(A).
\end{aligned}
\end{equation}
In Proof of the \cref{thm:conn_mcmc} we show that detailed balance conditions for continuous transitions imply a Poisson equation:
\begin{equation}
\begin{aligned}
F(I - Q_F)(A) = Z p_0(A),
\end{aligned}
\end{equation}
for a measurable set $A \subseteq \mathbb{R}^d$. By substituting $F_{\text{out}}=F$ and $F_{\text{in}}=(F_{\text{out}}-R)Q_F + Zp_0$, then
\begin{equation}
    (Z p_0 - R Q_F)(A) = F_{\text{in}}(A) - F_{\text{out}}Q_F(A).
\end{equation}
The left hand side is exactly the unnormalized term in the series of the measure $U$. Therefore, we can interpret $U(A)$ as an accumulated discrepancy between injected flow $F_{\text{in}}(A)$ and removed flow $F_{\text{out}}Q_F(A)$, when we applied a kernel $Q_F$.

\section{Proofs of results in the main text}
\label{sec:prooves}
\subsection{Derivation of flow representations}
\label{proof:derive_flow_from_db}
In this section, we derive key relations from the detailed balance conditions, which implies flow-matching as shown in \citep{lahlou2023theory}. Since this implication does not depend on acyclicity, it remains valid in our non-acyclic setting. Moreover, since all relevant measures and kernels in our framework admit densities with respect to the chosen reference measure, we may express these conditions in density form. Applying detailed balance for transitions from \(s_0\) to \(s \in \mathbb{R}^d\), yields
\begin{equation}
    f(s_0)\, p_F(s \mid s_0) = f(s)\, p_B(s_0 \mid s).
\end{equation}
Substituting \(p_F(s \mid s_0)=p_0(s)\), \(f(s_0)=Z\), and \(p_B(s_0 \mid s)=d_B(s)\), we obtain
\begin{equation}
\label{eq:flow_p0_clf}
    f(s) = \frac{Z\, p_0(s)}{d_B(s)}.
\end{equation}
Applying detailed balance to transitions from \(s \in \mathbb{R}^d\) to \(s_f\), we obtain
\begin{equation}
f(s)\, p_F(s_f \mid s) = f(s_f)\, p_B(s \mid s_f).
\end{equation}
Substituting \(p_F(s_f \mid s)=d_F(s)\), \(f(s_f)=Z\), and \(p_B(s \mid s_f)=r(s)/Z\), we arrive at the alternative representation
\begin{equation}
f(s) = \frac{r(s)}{d_F(s)}.
\end{equation}
Therefore, we have the following connection between classifiers:
\begin{equation}
d_F(s) = \frac{\pi(s)}{p_0(s)} d_B(s).
\end{equation}
\subsection{Proof of \cref{prop:flow_yiels_fm}.}
\label{proof:flow_yiels_fm}
From \cref{def:flow}, we have
\begin{equation}
\begin{aligned}
    &F(\{s_0\}) = F(\{s_f\})=Z,\\
    &F(A) = F(\{ s_0 \}) \sum_{n=0}^{\infty} P^n_F(s_0, A), 
    \qquad A \subseteq \bar{\mathcal{S}} \setminus \{ s_0,s_f\}.
\end{aligned}
\end{equation}
We aim to prove the flow-matching condition \eqref{eq:fm}, namely that for any measurable set \(A \subseteq \bar{\mathcal{S}} \setminus \{s_0\}\),
\begin{equation}
    F(A) = \int_{\bar{\mathcal{S}} \setminus \{ s_f\}} F(\mathrm ds)\, P_F(s, A).
\end{equation}
Case \(A \subseteq \mathbb{R}^d\).
Using the definition of \(F\), we obtain
\begin{equation}
\begin{aligned}
\int_{\bar{\mathcal{S}} \setminus \{ s_f \}} F(\mathrm ds)\, P_F(s, A)
&=
F(\{s_0\}) 
\int_{\bar{\mathcal{S}} \setminus \{ s_f \}} 
\sum_{n=0}^{\infty} P^n_F(s_0, \mathrm ds)\, P_F(s, A) \\
&\stackrel{(a)}{=}
F(\{s_0\}) 
\sum_{n=0}^{\infty} 
\int_{\bar{\mathcal{S}} \setminus \{ s_f \}} 
P^n_F(s_0, \mathrm ds)\, P_F(s, A) \\
&\stackrel{(b)}{=}
F(\{s_0\}) 
\sum_{n=0}^{\infty} P^{n+1}_F(s_0, A)\\
&=F(A) - F(\{s_0\}) \delta_{s_0}(A)\\ 
&=F(A).
\end{aligned}
\end{equation}
In step (a), we interchange the integral and the series using Tonelli's theorem. In step (b), we apply the Chapman--Kolmogorov relation:
\begin{equation}
P^{n+1}_F(s_0, A)
=
\int_{\bar{\mathcal{S}} \setminus \{ s_f\}} 
P^n_F(s_0, \mathrm ds)\, P_F(s, A),
\end{equation} because for \(A \subseteq \mathbb{R}^d\), we have \(P_F(s_f,A)=0\).
In the last step, noting that \(P_F^0(s_0,A)=\delta_{s_0}(A)=0\) for \(A \subseteq \mathbb{R}^d\) concludes the claim for this case.

Case \(A = \{s_f\}\).
In this case, the absorbing nature of \(s_f\) implies
\begin{equation}
\begin{aligned}
P^{n+1}_F(s_0, \{ s_f \})
&=
\int_{\bar{\mathcal{S}}} 
P^n_F(s_0, \mathrm ds)\, P_F(s, \{s_f\})\\
&= \int_{\bar{\mathcal{S}} \setminus \{ s_f\}} 
P^n_F(s_0, \mathrm ds)\, P_F(s, \{ s_f\})
+
P^n_F(s_0, \{ s_f\}),
\end{aligned}
\end{equation}
since \(P_F(s_f,\{s_f\})=1\). Repeating the steps from the previous case and using the above equation, yields:
\begin{equation}
\begin{aligned}
\int_{\bar{\mathcal{S}} \setminus \{ s_f \}} F(\mathrm ds)\, P_F(s, \{ s_f\})
&=
F(\{s_0\}) 
\sum_{n=0}^{\infty} 
\bigl(P^{n+1}_F(s_0, \{s_f\}) - P^n_F(s_0, \{s_f\})\bigr) \\
&=F(\{s_0\}) \bigg(\delta_{s_f}(\{ s_f \}) - \delta_{s_0}(\{ s_f \})\bigg)\\ &= F(\{s_0\}) = F(\{s_f\}),
\end{aligned}
\end{equation}
where under \cref{ass:finite_exp_len}, we have \(P^n_F(s_0,\cdot)\to \delta_{s_f}\). Combining these two cases, we conclude that the flow-matching conditions hold for all measurable sets \(A \subseteq \bar{\mathcal{S}} \setminus \{s_0\} \).
\qed
\subsection{Proof of \cref{prop:exp_traj_len_eq_flow}.}
\label{proof:exp_traj_len_eq_flow}
\begin{equation}
\begin{aligned}
    \mathbb{E}_{\tau \sim P}[n_{\tau}] = \mathbb{E}_{\tau \sim P}\bigg[\sum_{n=0}^{\infty}\mathds{1}\{ s_n \in \mathcal{S}\}\bigg] = \sum_{n=0}^{\infty} \mathbb P(s_n \in \mathcal{S}) = \sum_{n=0}^{\infty} P^n_F(s_0, \mathcal{S}) = \frac{F(\mathcal S)}{F(\{ s_0\})}.
\end{aligned}
\end{equation}
Note that under \cref{ass:finite_exp_len} the series above converges.
\qed
\subsection{Proof of \cref{thm:conn_mcmc} and \cref{cor:conn_mcmc}}
\label{proof:conn_mcmc}
\paragraph{Proof of \cref{thm:conn_mcmc}.}
From the detailed balance conditions at the source and sink states, we obtain for $s \in \mathbb{R}^d$ that:
\begin{equation}
\begin{aligned}
\label{eq:boundary_db}
f(s)d_B(s)&=Z\,p_0(s),\\
f(s)d_F(s)&=r(s).
\end{aligned}
\end{equation}
Now consider the detailed balance relation in the interior for \(s,s'\in\mathbb R^d\):
\begin{equation}
f(s)\bigl(1-d_F(s)\bigr)q_F(s'\mid s)
=
f(s')\bigl(1-d_B(s')\bigr)q_B(s\mid s').
\end{equation}
Integrating both sides with respect to \(s\), and using that \(q_B(\cdot\mid s')\) is a probability density, yields:
\begin{equation}
\int_{\mathbb{R}^d} f(s)\bigl(1-d_F(s)\bigr)q_F(s'\mid s)\,ds
=
f(s')\bigl(1-d_B(s')\bigr).
\end{equation}
Expanding both sides and using \eqref{eq:boundary_db}, gives the following:
\begin{equation}
\int_{\mathbb{R}^d} f(s)q_F(s'\mid s)\,ds
-
\int_{\mathbb{R}^d} r(s)q_F(s'\mid s)\,ds
=
f(s')-Z\,p_0(s').
\end{equation}
For the purpose of further proof, it will be convenient to write the above equation in a measure-kernel form. Define \(V(A) := F(A) / Z\) for \(A \subseteq \mathbb{R}^d\), and let \(\pi\), \(p_0\) and $\pi_Q$ denote the measures of the corresponding densities. Dividing the above equation by \(Z\), we obtain the Poisson equation:
\begin{equation}
\label{eq:pois}
V(I-Q_F)=p_0-\pi Q_F.
\end{equation}
Since \(Q_F\) is uniformly geometrically ergodic with a unique stationary measure \(\pi_Q\), there exist constants $C < \infty$ and $\rho \in (0,1)$ such that
\begin{equation}
\label{eq:geom_erg_tv}
    \| \nu \, Q^n_F - \pi_Q \|_{\text{TV}} \le C \rho^n,
\end{equation}
for all $n \in \mathbb{N}$ and any probability measure $\nu$, where $\| \cdot \|_{\text{TV}}$ is a total variation norm. Define a measure
\begin{equation}
U := \sum_{n=0}^{\infty} (p_0 Q^n_F - \pi Q^{n+1}_F),
\end{equation}
where the series converges absolutely in total variation using \eqref{eq:geom_erg_tv} and a triangular inequality. Then by telescoping
\begin{equation}
U - U Q_F =
\sum_{n=0}^{\infty}(p_0Q_F^n-\pi Q_F^{n+1})
-
\sum_{n=0}^{\infty}(p_0Q_F^{n+1}-\pi Q_F^{n+2})
=
p_0-\pi Q_F,
\end{equation}
which means that a measure $U$ is a solution to a Poisson equation \eqref{eq:pois}. Suppose that \(V\) is another solution to \eqref{eq:pois}, then
\begin{equation}
(V - U)(I - Q_F) = 0,
\end{equation}
which implies that \(V - U\) is an invariant measure for the kernel \(Q_F\). By the assumption that \(Q_F\) admits a unique stationary measure \(\pi_Q\), it follows that \(V - U\) lies in the one-dimensional subspace spanned by \(\pi_Q\). Therefore, all solutions of \eqref{eq:pois} are of the form
\begin{equation}
V = \sum_{n=0}^{\infty} \bigl(p_0 Q_F^n - \pi Q_F^{n+1}\bigr) + n_Q \, \pi_Q, \qquad n_Q \in \mathbb{R}.
\end{equation}
Since \(q_F(\cdot \mid s)\) is a transition density, the pushforward of any absolutely continuous measure remains absolutely continuous. Hence \(V\) admits a density of the form
\begin{equation}
v(s)=u(s)+n_Q\,\pi_Q(s),
\qquad n_Q\in\mathbb R,
\end{equation}
where by $u(s)$ we denote the corresponding density of a measure $U$.
Using the connection of a flow and a forward classifier in \eqref{eq:boundary_db}, we obtain
\begin{equation}
d_F(s)=\frac{r(s)}{f(s)}=\frac{\pi(s)}{v(s)}
=\frac{\pi(s)}{u(s)+n_Q\,\pi_Q(s)}.
\end{equation}
Similarly for the backward classifier from \eqref{eq:flow_reward_clf}
\begin{equation}
d_B(s) = \frac{p_0(s)}{\pi(s)} d_F(s) = \frac{p_0(s)}{u(s)+n_Q\,\pi_Q(s)}.
\end{equation}
Thus, in order for both classifiers to define valid probabilities, it is necessary and sufficient that
\begin{equation}
u(s)+n_Q\,\pi_Q(s)\ge \max\{\pi(s),p_0(s)\},
\qquad \forall s\in\mathbb R^d.
\end{equation}
The admissible values of \(n_Q\) are
\begin{equation}
n_Q \ge n^{*}_Q=
\sup_{s\in\mathbb R^d}
\frac{\max\{\pi(s),p_0(s)\}-u(s)}{\pi_Q(s)}.
\end{equation}
Finally, from the balance for the interior transitions
\begin{equation}
\label{eq:q_b_set}
    q_B(s \mid s') = \frac{(v(s) - \pi(s)) q_F(s' \mid s)}{v(s') - p_0(s')}.
\end{equation}
If the detailed balance conditions are satisfied and \cref{ass:finite_exp_len} holds, it follows that $n_Q$ and $n^{*}_Q$ are finite.

Conversely, given a solution $V$ to \eqref{eq:pois} with admissible values of $n_Q$, detailed balance conditions for a source-to-continuous and continuous-to-sink transitions are satisfied by setting $d_F(s) := \pi(s) / v(s)$ and $d_B(s) := p_0(s) / v(s)$. From \eqref{eq:q_b_set} balance conditions for the interior transitions are satisfied and by integrating over $\mathbb{R}^d$ we conclude that $q_B$ defines a valid transition density
\begin{equation}
\int_{\mathbb{R}^d} q_B(s \mid s') \mathrm ds = \frac{\int_{\mathbb{R}^d} (v(s) - \pi(s)) q_F(s' \mid s) \mathrm ds}{v(s') - p_0(s')} = \frac{v(s') - p_0(s')}{v(s') - p_0(s')} = 1,
\end{equation}
where we used that $(V-\pi)Q_F = V - p_0$, since $V$ is a solution to \eqref{eq:pois}. Hence, from $V$ we can uniquely recover all other objects to satisfy detailed balance conditions.
\qed

\paragraph{Proof of \cref{cor:conn_mcmc}.}
Notice that from \cref{prop:exp_traj_len_eq_flow}
\begin{equation}
\mathbb{E}[n_{\tau}] = \frac{F(\mathcal S)}{F(\{ s_0 \})} = V(\mathcal S) = U(\mathcal S) + n_Q \pi_Q(\mathcal S) = n_Q,
\end{equation}
because we are dealing with probability measures and $\mathcal S = \mathbb{R}^d$. Finally, minimizing the total flow is equivalent to minimizing
\begin{equation}
\int_{\mathbb{R}^d} f(s)\,ds = Z \int_{\mathbb{R}^d} v(s)\,ds =
Z\int_{\mathbb{R}^d} \bigl(u(s)+n_Q\,\pi_Q(s)\bigr)\,ds.
\end{equation}
Since \(\pi_Q\) is a probability density, this quantity is affine and strictly increasing in \(n_Q\). Therefore, the minimum is attained at the smallest admissible value of \(n_Q = n^{*}_Q\).

\end{document}